\documentclass[preprint,12pt]{elsarticle}

\usepackage{graphicx}%
\usepackage{multirow}%
\usepackage{amsmath,amssymb,amsfonts}%
\usepackage{amsthm}%
\usepackage{mathrsfs}%
\usepackage[title]{appendix}%
\usepackage{textcomp}%
\usepackage{manyfoot}%
\usepackage{booktabs}%
\usepackage[linesnumbered,lined,boxed,commentsnumbered]{algorithm2e}
\usepackage{algorithmicx}%
\usepackage{algpseudocode}%
\usepackage{listings}%
\usepackage{array}
\usepackage{subfigure}
\usepackage{stfloats}
\usepackage{url}
\usepackage{verbatim}
\usepackage{bbm}
\usepackage[table,xcdraw]{xcolor}
\usepackage{tabularx}
\usepackage{subfloat}
\usepackage{subcaption}
\usepackage{amssymb}
\usepackage{caption}
\usepackage{bbding}
\definecolor{mycustompurple}{RGB}{154, 36, 79} 
\usepackage[utf8]{inputenc}
\usepackage{xcolor} 
\usepackage{hyperref} 

\hypersetup{
    colorlinks=true,            
    linkcolor=red,              
    citecolor=green,            
    filecolor=mycustompurple,   
    urlcolor=magenta            
}

\journal{NEUROCOMPUTING}

\begin{document}

\begin{frontmatter}



\title{CRMSP: A Semi-supervised Approach for Key Information Extraction with Class-Rebalancing and Merged Semantic Pseudo-Labeling}



\author{Qi Zhang, Yonghong Song, Pengcheng Guo, Yangyang Hui}

\affiliation{organization={School of Software Engineering, Xi'an Jiaotong University},
            addressline={No.28, Xianning West Road}, 
            city={Xi'an City},
            postcode={710049}, 
            state={Shaanxi Province},
            country={China}}

\begin{abstract}
There is a growing demand in the field of KIE (Key Information Extraction) to apply semi-supervised learning to save manpower and costs, as training document data using fully-supervised methods requires labor-intensive manual annotation. The main challenges of applying SSL in the KIE are (1) underestimation of the confidence of tail classes in the long-tailed distribution and (2) difficulty in achieving intra-class compactness and inter-class separability of tail features. To address these challenges, we propose a novel semi-supervised approach for KIE with Class-Rebalancing and Merged Semantic Pseudo-Labeling (CRMSP). Firstly, the Class-Rebalancing Pseudo-Labeling (CRP) module introduces a reweighting factor to rebalance pseudo-labels, increasing attention to tail classes. Secondly, we propose the Merged Semantic Pseudo-Labeling (MSP) module to cluster tail features of unlabeled data by assigning samples to Merged Prototypes (MP). Additionally, we designed a new contrastive loss specifically for MSP. Extensive experimental results on three well-known benchmarks demonstrate that CRMSP achieves state-of-the-art performance. Remarkably, CRMSP achieves 3.24\% f1-score improvement over state-of-the-art on the CORD.
\end{abstract}

\begin{graphicalabstract}
\end{graphicalabstract}

\begin{highlights}
\item We propose a semi-supervised approach for KIE with Class-Rebalancing and Merged Semantic Pseudo-Labeling (CRMSP), utilizing a large number of unlabeled documents, significantly reducing the annotation costs, and improving the generalizability of the model.
\item To solve the problem of underestimation on the confidence of tail classes in the long-tailed distribution, we proposed the Class-Rebalancing Pseudo-Labeling (CRP) module.
\item We propose the Merged Semantic Pseudo-Labeling (MSP) module to fickle the difficulty in achieving intra-class compactness and inter-class separability of tail classes in unlabeled feature space.
\end{highlights}


\end{frontmatter}


\section{Introduction}
\label{introduction}

Key Information Extraction (KIE) as the downstream task of Optical Character Recognition (OCR) is the process of extracting structured information from documents.
KIE generally includes tasks such as named entity recognition and relation extraction, structured information extraction, and document classification. KIE has various applications in real-life scenarios, including bill processing, medical record handling, contract analysis, and resume processing. KIE is a challenging task since documents involve different types of information, including images, text, and layout. Recently, many multimodal pre-trained methods  \cite{xu2020layoutlmv2,huang2022layoutlmv3,guo2023dcmai} for KIE have been proposed to fickle this problem. However, these multimodal pre-trained methods require annotation for multiple types of information, which further increases time and manpower costs.


Semi-supervised learning (SSL) \cite{zhu2005semi} tackles situations with limited labeled and abundant unlabeled data \cite{sohn2020fixmatch,xu2021dash,zhang2021flexmatch,zheng2022simmatch,wang2022freematch}, bridging the gap between supervised and unsupervised learning for enhanced model performance. Existing SSL approaches \cite{sohn2020fixmatch,xie2020unsupervised} are to perform consistency regularization between weakly and strongly augmented views of unlabeled data based on the pseudo-labels predicted by the model as targets, thereby mitigating the model's sensitivity to small variations in similar samples within the input space. The performance of these SSL methods based on consistency regularization depends on whether the classes are balanced and the intra-class compactness and inter-class separability of the model in the feature space.
\begin{figure}[t]
    \begin{center}
        \includegraphics[width=0.8\textwidth]{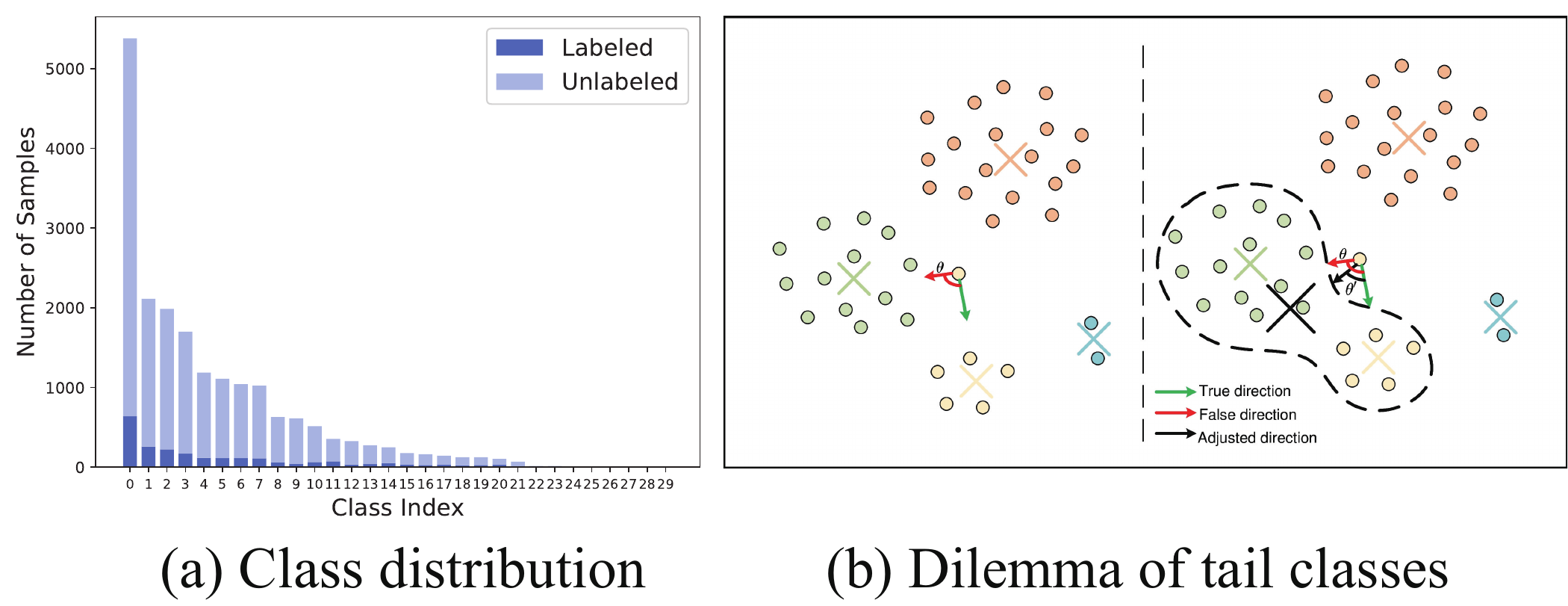}
    \end{center}
    \caption{Analysis on class distribution, and the dilemma of tail classes.
    }
    \label{problem}
\end{figure}
\begin{figure}[t]
    \centering
    \begin{minipage}[b]{0.32\linewidth}
      \centering
      \centerline{\includegraphics[width=\textwidth]{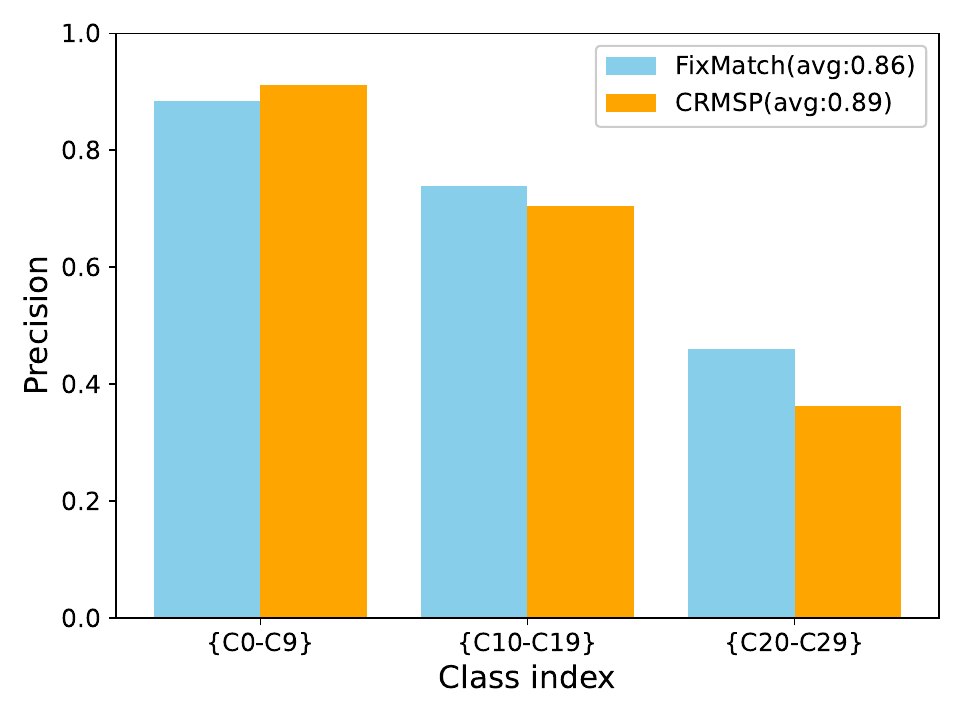}}
      \centerline{(a) Precision of PL}\medskip
    \end{minipage}
    \begin{minipage}[b]{0.32\linewidth}
      \centering
      \centerline{\includegraphics[width=\textwidth]{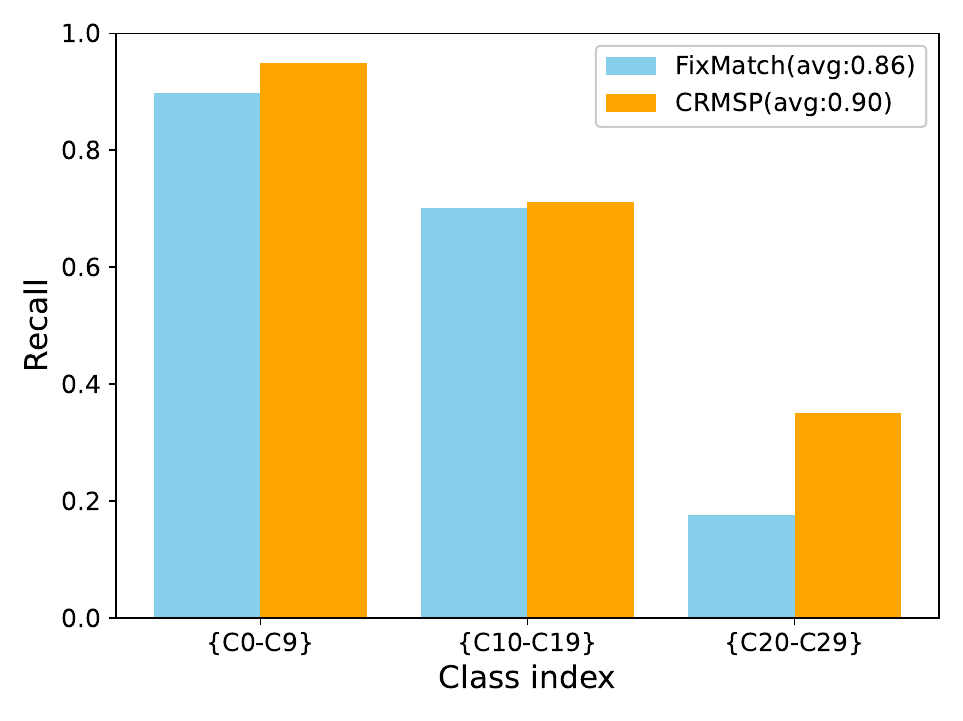}}
      \centerline{(b) Recall of PL}\medskip
    \end{minipage}
    \hfill
    \begin{minipage}[b]{0.32\linewidth}
      \centering
      \centerline{\includegraphics[width=\textwidth]{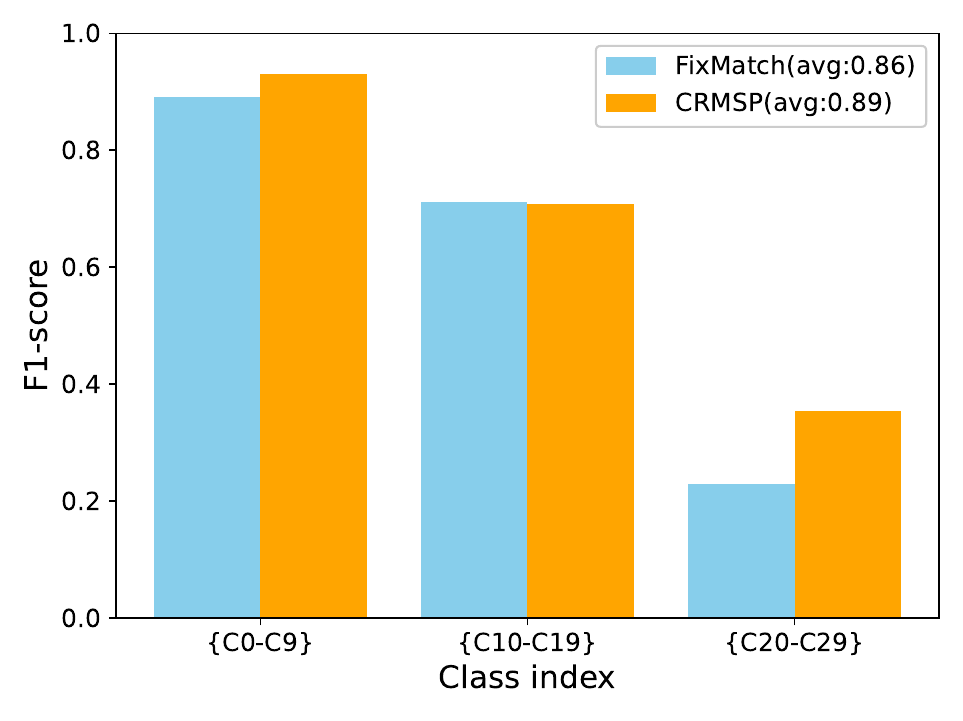}}
      \centerline{(c) F1-score of PL}\medskip
    \end{minipage}
    \caption{Comparison of precision, recall and f1-score of pseudo-labels generated by FixMatch and CRMSP. "PL" represents Pseudo-Labels.
    }
    \label{figure1}
\end{figure}

Specifically, the first one is that the confidence of tail classes in the long-tailed distribution \cite{zhang2023deep} is underestimated, leading to the model exhibiting higher confidence in predicting samples from the head classes. As shown in Fig. \ref{problem}(a), both labeled and unlabeled data exhibit a long-tailed distribution. This phenomenon implies that pseudo-labels are more likely to belong to head classes with higher probabilities and less likely to belong to tail classes with lower probabilities. As the number of iterations increases, this imbalance in the long-tailed distribution tends to worsen.

Secondly, it is hard to achieve intra-class compactness and inter-class separability of tail classes in unlabeled feature space \cite{huang2023semi}. To help the model learn richer representations, prototypes \cite{caron2020unsupervised,oh2022daso} are commonly introduced into the model. Each prototype can be seen as the representative features of a specific class of samples. The model calculates semantic pseudo-labels based on these prototypes. Following this, consistency regularization is applied to learn and enhance features. As shown in Fig. \ref{problem}(b), when classifying yellow unlabeled samples of tail classes, the sample points are closer to the "green" prototype than to "yellow". This results in "yellow" sample points being pushed away from the true direction and towards the wrong direction in the feature space. Therefore, the existence of the deviation angle $\theta$ causes semantic pseudo-labels to be biased to head classes over the tail classes, and this imbalance will increase as training progresses, leading to a degradation in model performance.

To address these challenges, we propose a semi-supervised approach for KIE with Class-Rebalancing and Merged Semantic Pseudo-Labeling (CRMSP). Firstly, to augment the model’s attention toward tail classes, we introduce the Class-Rebalancing Pseudo-Labeling (CRP) module that enhances the weight of pseudo-labels of tail classes and reduces the weight of pseudo-labels of head classes with a reweighting factor. It improves the recall of tail classes compared to the classical SSL method FixMatch \cite{sohn2020fixmatch} while maintaining a high precision of head classes, resulting in a higher f1-score, as shown in Fig. \ref{figure1}. 

Secondly, to enhance the intra-class compactness and separability from other classes of tail classes, we propose the Merged Semantic Pseudo-Labeling (MSP) module. This module sorts the generated pseudo-labels in descending order, identifies the Top-$K$ classes, and aggregates the features of these $K$ classes in the memory bank into a super-class. The merged prototype (MP) is then calculated. By utilizing the clustering of merged prototypes, the semantic pseudo-labels generated in this way push the features of tail samples closer to the prototype of the super-class, rather than pushing the features closer to prototypes of head classes.

Additionally, we designed a new contrastive loss specifically for the merged semantic pseudo-labels, whose effectiveness is demonstrated in Table \ref{table3}. Extensive experimental results indicate that the MSP module improves the performance of tail classes.

To the best of our knowledge, CRMSP is the first semi-supervised learning method in the field of KIE. Based on a multi-modal model, CRMSP has designed a semi-supervised learning method that fully utilizes text, image, and layout information, which is different from previous semi-supervised methods that only utilize image or text information from CV or NLP models. The main contributions are summarized as follows:

\begin{itemize}
\item[$\bullet$] We propose a semi-supervised approach for KIE with Class-Rebalancing and Merged Semantic Pseudo-Labeling (CRMSP), utilizing a large number of unlabeled documents, significantly reducing the annotation costs, and improving the generalizability of the model.
\item[$\bullet$] To solve the problem of underestimation of the confidence of tail classes in the long-tailed distribution, we proposed the Class-Rebalancing Pseudo-Labeling (CRP) module.
\item[$\bullet$] We propose the Merged Semantic Pseudo-Labeling (MSP) module to fickle the difficulty in achieving intra-class compactness and inter-class separability of tail classes in unlabeled feature space.
\end{itemize}

\section{Related Work}
\label{related-work}

\subsection{Key information extraction} Transformer-based pre-training has demonstrated success across various KIE tasks, where extensive unlabeled document datasets are leveraged for model pre-training, preceding fine-tuning on downstream tasks. Numerous existing frameworks \cite{xu2020layoutlm,xu2020layoutlmv2,huang2022layoutlmv3,guo2023dcmai} have investigated pre-training approaches on documents. LayoutLM \cite{xu2020layoutlm} achieved significant improvements in various document understanding tasks by jointly pre-training text and layout. LayoutLMv2 \cite{xu2020layoutlmv2} greatly enhanced the model's image understanding capability by integrating visual feature information into the pre-training process. LayoutLMv3 \cite{huang2022layoutlmv3} overcame the differences between text and image in pre-training objectives and promoted multi-modal representation learning. Our approach utilizes these multi-modal models based on image, text, and layout as encoders, extending the scope of semi-supervised methods to the KIE domain.

\subsection{Semi-supervised learning} SSL is a learning approach focused on building models that leverage both labeled and unlabeled data. While unlabeled data is crucial for SSL, generating pseudo-labels from model predictions remains a challenge. Existing approaches, including pseudo-labeling \cite{xie2020self}, consistency regularization \cite{tarvainen2017mean,laine2016temporal}, generative methods \cite{fang2020triple,liu2020catgan} and hybrid methods \cite{berthelot2019remixmatch,sohn2020fixmatch,xu2021dash,zhang2021flexmatch,zheng2022simmatch,wang2022freematch}. However, pseudo-labels can introduce bias, particularly in the presence of imbalanced data, adversely affecting model performance. To mitigate this issue, previous works have explored various strategies such as threshold adjustment \cite{xu2021dash,zhang2021flexmatch,wang2022freematch}, incorporating additional classifiers \cite{kuo2020featmatch,oh2022daso}. However, designing dynamic thresholds is complex and computationally intensive. In our work, we directly incorporate an additional branch for semantic pseudo-label classification, which effectively promotes intra-class compactness and inter-class separability for imbalanced classes, without the need for designing complex dynamic threshold strategies.

\subsection{Imbalanced learning} Class-imbalanced supervised learning is of great interest both in theory and in practice. Recent works include resampling \cite{chawla2002smote,wei2021crest} and reweighting \cite{hyun2020class} which rebalance the contribution of each class, while others focus on reweighting the given loss function by a factor inversely proportional to the sampling frequency in a class-wise manner. \cite{hyun2020class} proposed a suppressed consistency loss to suppress the loss on minority classes.\cite{kim2020distribution} proposed Distribution Aligning Refinery (DARP) to refine pseudo-labels for SSL under assuming class-imbalanced training distributions. CReST proposed a re-sampling method to iteratively refine the model by supplementing the labeled set with high-quality pseudo-labels, where minority classes are updated more aggressively than majority classes. DASO adaptively blends the linear and semantic pseudo-labels within each class to mitigate the overall bias across the class for imbalanced semi-supervised learning. In our work, we alleviate the class-imbalanced problem by directly rebalancing pseudo-labels according to distributions between head and tail classes instead of designing complicated reweighting losses.

\section{Proposed Method}
\label{method}

\subsection{Preliminaries}
For a $C$-class semi-supervised classification problem, let $\mathcal{X}=\{(x_b,y_b)\}_{b=1}^B$ be a batch of $B$ labeled samples, where $x_b$ are the training samples and $y_b$ are the ground-truth, $y_b\in\mathcal{Y}=\{1,\ldots,C\}$. Meanwhile, let $\mathcal{U}=\{u_{b}\}_{b=1}^{\mu B}$ be a batch of $\mu B$ unlabeled samples, where the hyperparameter $\mu$ is used to control the batch size of unlabeled samples. Note that the underlying ground truth $\hat{y}$ of unlabeled data may be different from labeled data, $\hat{y}\in\mathcal{Y}$, $\mathcal{Y}=\{1,\ldots,C\}$.

\begin{figure*}[t]
    \begin{center}
        \includegraphics[width=\textwidth, trim=20 20 20 20, clip]{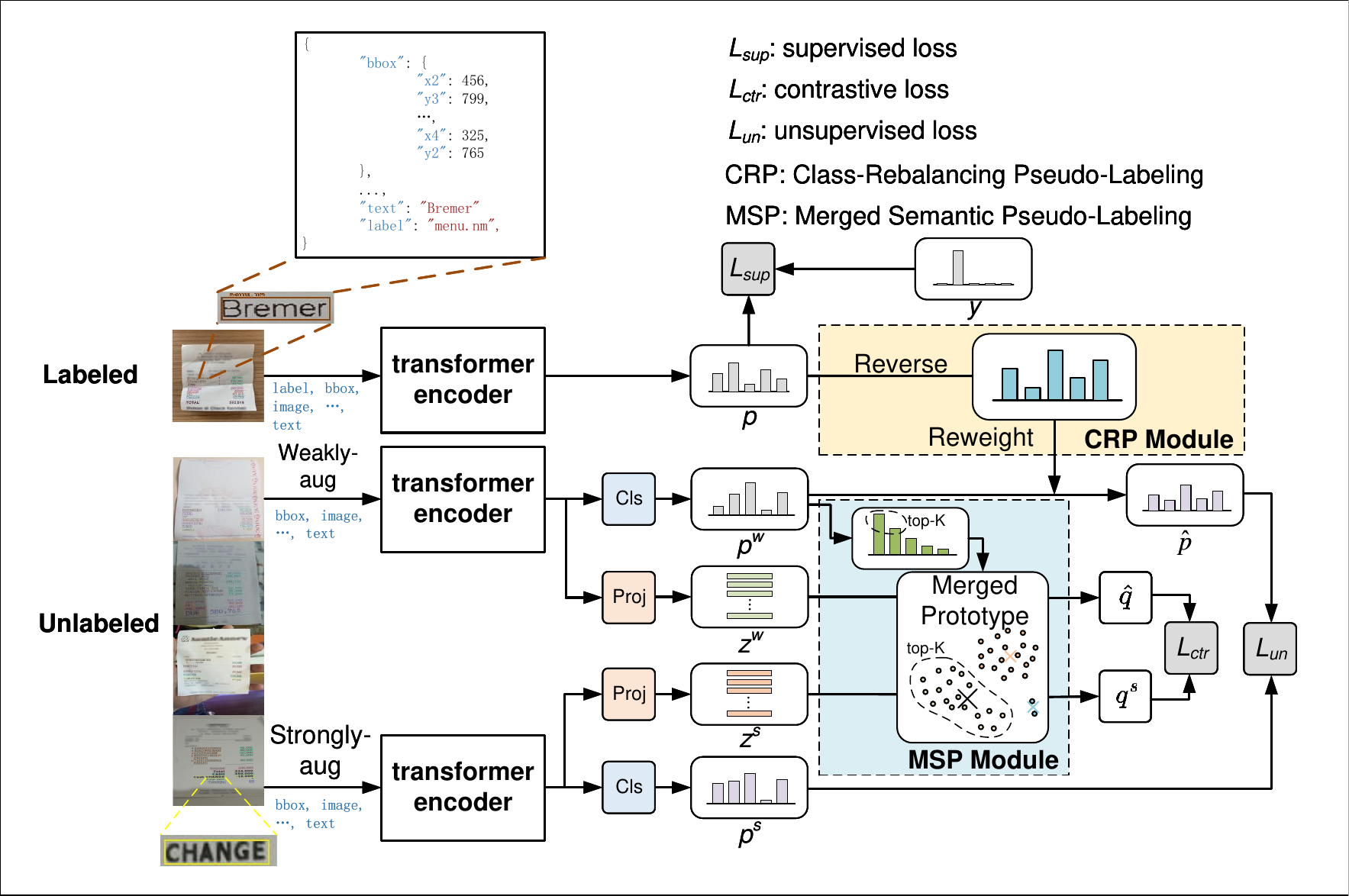}
    \end{center}
    \caption{Framework of the proposed Class-Rebalancing and Merged Semantic Pseudo-Labeling (CRMSP). Labeled and unlabeled samples are from the training data mini-batch. 
    }
    \label{figure2}
\end{figure*}

For the labeled data, the input $x_b$ is paired with the label $y_b$ to train the base model $f(\cdot)$ through calculating supervised loss $\mathcal{L}_{sup}$, generating features $z_b$. For the unlabeled data, unlabeled samples are sent to the base model $f(\cdot)$ as inputs after weak augmentation $\mathcal{A}_w$ and strong augmentation $\mathcal{A}_s$. Both are followed by a classification head $h(\cdot)$ and a projection head $g(\cdot)$ to get $p^{w}=h\circ f(\mathcal{A}_w(u))$, $z^{w}=g\circ f(\mathcal{A}_w(u))$, $p^{s}=h\circ f(\mathcal{A}_s(u))$ and $z^{s}=g\circ f(\mathcal{A}_s(u))$. The Class-Rebalancing Pseudo-labeling module is employed to alleviate the imbalance problem of pseudo-labels. The rebalanced pseudo-labels $\hat{p}\in\mathbb{R}^{C}$ are then assigned to calculate the unsupervised loss $\mathcal{L}_{un}$. The Merged Semantic Pseudo-Labeling module generates merged semantic pseudo-labels of unlabeled features with the Merged Prototypes $\mathbf{\hat{C}}$, which is used to compute the contrastive loss $\mathcal{L}_{ctr}$. The overall framework is shown in Fig. \ref{figure2}.

\subsection{Class-Rebalancing Pseudo-Labeling}
\label{sec:rebalance}
Due to the smaller sample size in the tail classes compared to the head classes, the model tends to generate lower confidence when predicting tail data. The approach in FixMatch \cite{sohn2020fixmatch}, which filters out samples based on a fixed threshold applied to the highest confidence, overlooks the numerical disadvantage of the tail data. Experiments show that the predictive distribution of labeled samples is generally positively correlated with the distribution of unlabeled samples. We estimate the approximate sample distribution by calculating the exponential moving average (EMA) of the model's confidence predictions for labeled data $p$, which we call $\tilde{p}$. Note that EMA preserves previous information through a weighted average, smoothing the process of updating data. At the $t$-th iteration, we compute $\tilde{p}_t$ as:

\begin{equation}
    \tilde{p}_t =
\begin{cases}
  \frac{1}{K}, & \text{if } t = 0, \\
  \lambda\tilde{p}_{t-1}+(1-\lambda)\frac{1}{B}\sum_{b=1}^{B}p_b,   & \text{otherwise},
\end{cases}
\end{equation}

where $\lambda$ is a smoothing factor. 

\subsubsection{Reweighting pseudo-labels} We observed that the pseudo-labels produced by the model are biased toward the head classes. To augment the model's attention towards tail classes, we introduced a reweighting factor $\beta$ that enhances the weight of tail classes while correspondingly reducing that of head classes.

We first perform the reverse operation on $\tilde{p}_t$ to obtain the reweighting factor $\beta$:
\begin{equation}
\label{sub:eq2}
\beta=Normalize(\mathcal{M}(\tilde{p}_t)),
\end{equation}
$\mathcal{M}$ is a monotonically decreasing mapping function (a minimum value not less than 0) that ensures a higher weight is assigned to classes with a smaller proportion in the predicted distribution, while classes with a larger proportion in the predicted distribution receive a lower weight. e.g., $\mathcal{M}(x)=1-x/T$ , where $T$ is a temperature hyperparameter, $Normalize(\cdot)$ is the normalized operation defined as $x'_i=x_i/ {\textstyle \sum_{j=1}^{n}} x_j,i\in(1,\ldots,n)$.

Then the model's confidence predictions $p_b^w$ for weakly-augmented data is multiplied by $\beta$ to get more balanced confidence predictions $p_{b}^{\prime}$ in a batch:

\begin{equation}{p_{b}^{\prime}}=Normalize({p_b^w}\times \beta)\end{equation}
The rebalanced pseudo-labels $\hat{p}_b$ are generated by $argmax(\cdot)$ in a batch:
\begin{equation}
\hat{p}_b=argmax({p_{b}^{\prime}}).
\end{equation}
By rebalancing the pseudo-labels, tail labels are more likely to be chosen when filtering pseudo-labels with a fixed threshold for prediction.

\subsection{Merged Semantic Pseudo-Labeling}

To obtain semantic pseudo-labels from a feature perspective, DASO \cite{oh2022daso} involves prototype clustering, which updates the dynamic memory bank with features and ground-truths of labeled data. However, due to the significantly smaller number of tail samples compared to head classes, the tail features in this memory bank are inherently limited. Consequently, the computed tail prototypes lack representation, and it is inappropriate to assume that all tail features are concentrated around this prototype. The semantic pseudo-labels are computed by merely comparing tail samples to this unrepresentative prototype push tail features close to prototypes of other classes, which is detrimental to achieving intra-class compactness and inter-class separability of tail classes.

\subsubsection{EMA Model} 

The basic assumption in SSL is the smoothness assumption: if two data points are close in high-density regions, their corresponding outputs should also be close. Mean Teacher \cite{tarvainen2017mean} utilizes this assumption by using unlabeled data. In practice, augmented samples are generated by adding small perturbations to the original samples, and they should have consistent predictions in both Teacher and Student models, achieved through consistency constraints. The Teacher model is essentially an EMA model of the Student. The EMA model provides guidance for updating the parameters of the base model. Therefore, their weights are tightly coupled. The parameter ($\theta'$) of the EMA model is updated with a weighted average of the current parameter ($\theta$) of the base model: 
\begin{equation}
\theta'_t=\alpha\cdot\theta'_{t-1}+(1-\alpha)\cdot\theta
\end{equation}
where $\alpha$ is a smoothing factor ($0<\alpha<1$).

\subsubsection{Merged Prototypes (MP) generation}
We first build a set of basic prototypes $\mathbf{C}=\{c_{i}\}_{i=1}^{C}$ from $\mathcal{X}$. The basic prototype $c_i$ for every class is efficiently calculated by averaging the feature representations in the dynamic memory bank $\mathbf{Q}=\{Q_{i}\}_{i=1}^{C}$, $Q_{i}=\{z_{j}\}_{j=1}^{maxsize}$, where $Q_{i}$ is a queue with a max size. We update $\mathbf{Q}$ every iteration by pushing new features $z_b$ and labels $y_b$ from a batch of labeled data.

Then we determine to construct the super-class for each batch. Based on a common understanding: for an unlabeled sample, if the confidences of several classes are close, their corresponding feature representations in the feature space are close. In such cases, we merge these top-$K$ proximate classes. We achieve this by sorting the confidence predictions $p_{b}^{w}$ obtained from the weak augmentation branch in descending order, resulting in the sorted confidence predictions $s_{b}^{w}$:
\begin{equation}
    s_{b}^{w} = \text{sort}(p_{b}^{w}, \text{descending}) =\{p_{\sigma(1)}^{w}, \ldots, p_{\sigma(K)}^{w}, \ldots, p_{\sigma(C)}^{w}\}
\end{equation}
where $\{\sigma(1),\ldots,\sigma(C)\}$ represents the order in which classes are arranged in descending order.

After obtaining this order, we merge the features corresponding to the top-$K$ classes $Q_{\sigma(K)}$ in the dynamic memory bank $\mathbf{C}$ to get the new dynamic memory bank is $\mathbf{\hat{Q}}=\{\hat{Q_{i}}\}_{i=1}^{N}$, where $N=C-K+1$:
\begin{equation}
\hat{Q_{i}}= 
\begin{cases}
    Q_{\sigma (1)}\cup Q_{\sigma (2)}\cup \ldots\cup Q_{\sigma (K)}, & \text{if } i = 1 \\
    Q_{\sigma (K+i-1)}, & \text{otherwise}
\end{cases}
\end{equation}

For Merged prototypes $\mathbf{\hat{C}}=\{\hat{c_{i}}\}_{i=1}^{N}$, each $\hat{c_{i}}$ is computed by taking the average of the features in the queue $\hat{Q_{i}}$.

\subsubsection{Semantic pseudo-labels} In a batch, the merged semantic predictions $q_{b}^{w}$ and $q_{b}^{s}$ of the super class space is computed from $z_{b}^{w}$, $z_{b}^{s}$ with the merged prototype $\mathbf{\hat{C}}$:
\begin{equation}
q_{b}^{w}=Sim(z_{b}^{w},\mathbf{\hat{C}})=\frac{<z_{b}^{w},\mathbf{\hat{C}}>}{\parallel z_{b}^{w}\parallel \parallel \mathbf{\hat{C}}\parallel } / T_{proto}
\end{equation}

\begin{equation}
q_{b}^{s}=Sim(z_{b}^{s},\mathbf{\hat{C}})=\frac{<z_{b}^{s},\mathbf{\hat{C}}>}{\parallel z_{b}^{s}\parallel \parallel \mathbf{\hat{C}}\parallel } / T_{proto}
\end{equation}
where $T_{proto}$ is a temperature hyperparameter, and $Sim(\cdot)$ denotes cosine similarity, $<\cdot,\cdot>$ represents the dot product operation, and $\parallel \cdot \parallel$ represents the L2 norm.

The merged semantic pseudo-labels $\hat{q}_b$ are generated by $argmax(\cdot)$ in a batch:
\begin{equation}
\hat{q}_b=argmax({q_{b}^{w}}).
\end{equation}

\subsection{Loss function} 
Following the SSL paradigm, the first two items are supervised loss $\mathcal{L}_{sup}$ and unsupervised loss $\mathcal{L}_{un}$, respectively. In addition, we include a contrastive loss $\mathcal{L}_{ctr}$ to compute the distance of two semantic similarities. The loss minimized by CRMSP is simply:
\begin{equation}
\mathcal{L}_{total}=\mathcal{L}_{sup}+\lambda_{un}\mathcal{L}_{un}+\lambda_{ctr}\mathcal{L}_{ctr},
\label{total-loss}
\end{equation}
\begin{equation}
\mathcal{L}_{sup}=\dfrac{1}{B}\sum_{b=1}^{B}\mathcal{H}(p_b,y_b),
\end{equation}

\begin{equation}
\mathcal{L}_{un}=\dfrac{1}{\mu B}\sum_{b=1}^{\mu B}\mathbbm{1}(max({p_{b}^{\prime}})\ge \tau )\mathcal{H}(\hat{p}_b,p_b^{s}),
\end{equation}

\begin{equation}
\mathcal{L}_{ctr}=\dfrac{1}{\mu B}\sum_{b=1}^{\mu B}\mathcal{H}(\hat{q_{b}},q^{s}_b).
\end{equation}
where $\lambda_{un}$ and $\lambda_{ctr}$ are fixed scalar hyperparameters denoting the relative weight of the unsupervised loss and contrastive loss, and $\mathbbm{1}$ is all-one vector, $\tau$ represents a fixed threshold. $\mathcal{H}$ represents cross-entropy.

\subsection{Pseudo Code}
Algorithm \ref{pseudo_code} presents the algorithm of the entire CRMSP during the training phase.

\begin{algorithm}

\KwData{A batch of labeled and unlabeled samples $\mathcal{X}=\{(x_b,y_b)\}_{b=1}^B$ and $\mathcal{U}=\{u_{b}\}_{b=1}^{\mu B}$, base model $f(\cdot)$, classification head and projection head: $h(\cdot)$ and $g(\cdot)$} 

\SetKwInOut{Input}{Input}
\SetKwInOut{Output}{Output}

\Input{$\lambda_{un}$: weight of the unsupervised loss, $\lambda_{ctr}$: weight of the contrastive loss, $\tau$: a fixed threshold}
\Output{Trained model parameters}

\BlankLine

\For{epoch in range(num\_epochs)}{
    \For{batch in unlabeled\_data\_batches}{

        \BlankLine
        \{\textbf{Supervised Learning with labeled data}\}\;
        $p_{b}=h\circ f(x_{b})$,
        $ \tilde{p}_t =
        \begin{cases}
          \frac{1}{K}, & \text{if } t = 0, \\
          \lambda\tilde{p}_{t-1}+(1-\lambda)\frac{1}{B}\sum_{b=1}^{B}p_b,   & \text{otherwise},
        \end{cases}$\;
        $\mathcal{L}_{sup}=\dfrac{1}{B}\sum_{b=1}^{B}\mathcal{H}(p_b,y_b)$\;

        \BlankLine
        \{\textbf{Unsupervised Learning with unlabeled data}\}\;
        $p_{b}^{w}=h\circ f(\mathcal{A}_w(u_{b}))$, $p_{b}^{s}=h\circ f(\mathcal{A}_s(u_{b}))$\;
        $\beta=Normalize(\mathcal{M}(\tilde{p}_t))$,  ${p_{b}^{\prime}}=Normalize({p_b^w}\times \beta)$\;
        
        $\hat{p}_b=argmax({p_{b}^{\prime}})$\;

        $\mathcal{L}_{un}=\dfrac{1}{\mu B}\sum_{b=1}^{\mu B}\mathbbm{1}(max({p_{b}^{\prime}})\ge \tau )\mathcal{H}(\hat{p}_b,p_b^{s})$\;
        
        \BlankLine
        \{\textbf{Contrastive Learning}\}\;
        $z_{b}^{w}=g\circ f(\mathcal{A}_w(u_{b}))$, $z_{b}^{s}=g\circ f(\mathcal{A}_s(u_{b}))$\;
        
        $\mathbf{Q}=\{Q_{i}\}_{i=1}^{C}$, $Q_{i} \leftarrow(z_b, y_b)$\;

        $\mathbf{C}=\{\hat{c_{i}}\}_{i=1}^{N}\leftarrow$ the average of the features $\mathbf{Q}$\;
        
        $s_{b}^{w} = \text{sort}(p_{b}^{w}, \text{descending}) =\{p_{\sigma(1)}^{w}, \ldots, p_{\sigma(K)}^{w}, \ldots, p_{\sigma(C)}^{w}\}$, $\{\sigma(1),\ldots,\sigma(C)\}$ represents the descending order\;

        $\mathbf{\hat{Q}}=\{\hat{Q_{i}}\}_{i=1}^{N}$, $\mathbf{\hat{C}}=\{\hat{c_{i}}\}_{i=1}^{N}\leftarrow$the average of the features $\mathbf{\hat{Q}}$\;
        $q_{b}^{w}=\frac{<z_{b}^{w},\mathbf{\hat{C}}>}{\parallel z_{b}^{w}\parallel \parallel \mathbf{\hat{C}}\parallel } / T_{proto}, q_{b}^{s}=\frac{<z_{b}^{s},\mathbf{\hat{C}}>}{\parallel z_{b}^{s}\parallel \parallel \mathbf{\hat{C}}\parallel } / T_{proto}$\;

        $\hat{q}_b=argmax({q_{b}^{w}})$\;
        
        $\mathcal{L}_{ctr}=\dfrac{1}{\mu B}\sum_{b=1}^{\mu B}\mathcal{H}(\hat{q_{b}},q^{s}_b)$\;

        \BlankLine
        $\mathcal{L}_{total}=\mathcal{L}_{sup}+\lambda_{un}\mathcal{L}_{un}+\lambda_{ctr}\mathcal{L}_{ctr}$\;
        
        update $f(\cdot)$, $h(\cdot)$, $g(\cdot)$ with AdamW to minimize $\mathcal{L}_{total}$\;
    }
}

\caption{CRMSP algorithm}
\label{pseudo_code}
\end{algorithm}

\section{Experiments}
\label{experiments}
\subsection{Datasets and Compared Methods}
\subsubsection{Datasets} FUNSD \cite{jaume2019funsd} is a comprehensive collection of real, fully annotated, scanned forms with 149 samples for training and 50 samples for testing. The documents are noisy and vary widely in appearance, making form understanding a challenging task. The proposed dataset can be used for various tasks, including text detection, optical character recognition, spatial layout analysis, and entity. CORD \cite{park2019cord} is typically utilized for receipt KIE, which consists of 800/100/100 receipts for training/validation/testing. The dataset consists of thousands of Indonesian receipts, which contain images and box/text annotations for OCR, and multi-level semantic labels for parsing. The proposed dataset can be used to address various OCR and parsing tasks.

We construct long-tailed versions of CIFAR10 (CIFAR10-LT), CIFAR100 (CIFAR100-LT), and STL10 (STL10-LT) separately. $N_k$ represents the number of examples in class $k$ for labeled data and unlabeled data. $\gamma_{l}$ and $\gamma_{u}$ are the imbalance ratios for labeled data and unlabeled data. The number of examples for each class except the head class is based on the formula $N_{k}=N_{1}\cdot\gamma_{l}^{-\frac{k-1}{K-1}}$. It is important to note that within each $N_{k}$, examples in class $k$ are arranged in descending order (i.e., $N_{1}\geq\cdots\geq N_{K}$).

\subsubsection{Compared Methods} Our approach is compared with both classical and imbalanced SSL methods. Classical SSL methods include FixMatch \cite{sohn2020fixmatch}, Dash \cite{xu2021dash}, FlexMatch \cite{zhang2021flexmatch}, SimMatch \cite{zheng2022simmatch} and FreeMatch \cite{wang2022freematch}. Imbalanced SSL methods include DARP \cite{kim2020distribution}, CReST \cite{wei2021crest} and DASO \cite{oh2022daso}.
\subsection{Experiments settings} 
Our experiments are conducted on NVIDIA Tesla V100 GPU. For KIE datasets, the split ratios (i.e., the proportions of labeled data) are 5\% and 10\%, and the batch size of labeled data is 4, $\mu$ is set to 1.0. To validate the effectiveness of our proposed SSL approach, we use Transformer-based models such as LayoutLMv2 and LayoutLMv3 as base models. For all methods, we employ the Adam optimizer with a fixed learning rate of 1e-5. we take precision, recall and f1-score as our evaluation metrics. Referring to \cite{wang2022usb}, we set weak augmentation to none and strong augmentation to random swap, which means randomly swapping two words in the sentence $n$ times. In training, temperature hyperparameter $T_{proto}$ is set to 1.0. The EMA model is used for testing with a momentum factor 0.999. For two unsupervised supervised losses, both $\lambda_{un}$ and $\lambda_{ctr}$ are set to 0.1, $\tau$ is set to 0.95. For both LayoutLMv2 and LayoutLMv3, $K$ was fine-tuned with a setting of 5.

For CV datasets, we train the CIFAR10-LT/CIFAR100-LT and STL10-LT on the Wide ResNet-28-2 \cite{kim2020distribution} with 1.5M parameters for 250$k$ iterations. The optimizer is SGD with a learning rate of 0.03 and a weight decay of 5e-4. We set the training epoch to 256 and the batch size to 64. The imbalance ratio includes the following settings: $\gamma=\gamma_{l}=\gamma_{u}=100$, $\gamma=\gamma_{l}=\gamma_{u}=10$ and $\gamma_{l}=10, \gamma_{u}: unknown$. The number of examples for the head class of labeled data $N_{1}$ is set to $\{500, 1500, 50, 150, 450\}$, and the number of examples for the head class of unlabeled data $M_{1}$ is set to $\{4000, 3000, 400, 300, 100k\}$.

\begin{table}[]
    \centering
    \resizebox{\textwidth}{!}{%
    \begin{tabular}{cccccccccccccccccc}
\hline
\multirow{3}{*}{Base Model} & \multirow{3}{*}{Method} & \multicolumn{6}{c}{FUNSD}                          & \multicolumn{6}{c}{CORD}                           \\ \cmidrule(r){3-8} \cmidrule(r){9-14}
                            &                         & \multicolumn{3}{c}{5\%} & \multicolumn{3}{c}{10\%} & \multicolumn{3}{c}{5\%} & \multicolumn{3}{c}{10\%} \\ \cmidrule(r){3-5} \cmidrule(r){6-8} \cmidrule(r){9-11} \cmidrule(r){12-14} 
                            &                         & P   & R   & F1    & P   & R   & F1      & P   & R   & F1     & P   & R   & F1    \\ \hline
    \multirow{6}{*}{LayoutLMv2}                     & Fully-supervised        & 77.91         & 81.38          & 79.61          & 77.91         & 81.38          & 79.61         & 95.81          & 95.02     & 95.41       & 95.81          & 95.02     & 95.41    \\ \cmidrule(r){2-14}
                                                    & FixMatch \cite{sohn2020fixmatch}    & 61.42          & 70.55     & 65.67      & 65.27          & 73.36     & 69.08        &  83.96    &  85.36       & \underline{84.66}    & 87.01     & 87.47        & 87.24     \\
                                                    & Dash \cite{xu2021dash}         & 62.71          & 74.51     & \textbf{68.10}    & 64.77          & 70.65       & 67.58   & 77.99    & 82.64    & 80.25      & 87.47       & 88.53 & 88.00         \\
                                                    & FlexMatch \cite{zhang2021flexmatch}                  & 59.14          & 75.66         & 66.39    & 62.85          & 72.40         & 67.29       & 84.60          & 84.60   & 84.60   & 86.84          & 87.17   & 87.01     \\
                                                    & SimMatch \cite{zheng2022simmatch}       &  63.29         & 70.85         & 66.86        & 70.06 & 77.72 & \underline{73.69} &  82.83         &  83.40         & 83.11        &  88.21   & 89.21 & 88.71    \\
                                                    & FreeMatch \cite{wang2022freematch}    & 64.06     & 70.65     & 67.19      & 68.32   & 74.46  & 71.26    & 82.79          & 84.60   & 83.69   & 88.81          & 89.28   & \underline{89.05}     \\
                                                    & CRMSP(Ours)             & 63.39 & 71.85  & \underline{67.36} & 71.49 & 78.63  & \textbf{74.89}  & 84.47 & 84.98 & \textbf{84.73} & 90.46 & 90.19 & \textbf{90.33}  \\ \hline
    \multirow{6}{*}{LayoutLMv3}                     & Fully-supervised        & 77.74          & 80.33          & 79.01        & 77.74          & 80.33          & 79.01             & 93.09      & 93.51          & 93.30        & 93.09      & 93.51          & 93.30   \\ \cmidrule(r){2-14}
                                                    & FixMatch \cite{sohn2020fixmatch}               & 57.40          & 67.46          &  62.02      &  62.74      &  73.87      &  67.85     &  82.73        &  83.17         & \underline{82.95}         & 86.72  & 87.25 
 & 86.98   \\
                                                    & Dash \cite{xu2021dash}                 &  53.53         &  63.34         &  58.02      & 
 61.91    &  67.41      & 64.54     &  82.11         & 83.47          &  82.78     & 
 83.77    &  86.87      & 85.29   \\
                                                    & FlexMatch \cite{zhang2021flexmatch}               & 60.04          & 70.54         & 64.87       & 70.85    & 73.77       &   \underline{72.28}       & 80.06          & 82.42          & 81.22      & 86.10  & 86.49  & 86.30    \\
                                                    & SimMatch \cite{zheng2022simmatch}                    &  64.78         &      66.68    & 65.72        & 69.40 & 77.40 & \textbf{73.18} & 82.07          & 82.57          & 82.32       &  87.72 & 87.85 & 87.78  \\
                                                    & FreeMatch \cite{wang2022freematch}                    & 60.37          & 72.88          & \underline{66.04}         &  67.17     &  71.14   & 
 69.10  &  82.06         &  82.87         &  82.46        &   87.08     &  89.06    & \underline{88.06}   \\
                                                    & CRMSP(Ours)    & 63.89 & 71.98 & \textbf{67.69}   & 67.68  &  74.91   & 71.12   & 82.41 & 84.15 & \textbf{83.27} & 91.51 & 91.09 & \textbf{91.30}\\ \hline
    \end{tabular}
    }
    \caption{Evaluation results with 5\% and 10\% of labeled training samples on three KIE benchmarks based on LayoutLMv2 and LayoutLMv3. The best result is in \textbf{bold}, the second-best result is in \underline{underline}.}
    \label{main-results2}
\end{table}

\subsection{Results}
\subsubsection{Results on FUNSD and CORD}
To validate the effectiveness of CRMSP, we perform experiments on FUNSD and CORD for the token classification task. Table \ref{main-results2} shows comparative results with 5\% and 10\% labeled samples based on LayoutLMv2 and LayoutLMv3. For all methods, we observe that the f1-score increases as the ratio of labeled data increases. In contrast, CRMSP improves the f1-score in the vast majority of experimental settings. Based on LayoutLMv3, it works particularly well on the CORD with 10\% labeled data and achieves 4.32\% and 3.24\% f1-score gain compared with FixMatch and suboptimal method FreeMatch, respectively. On the FUNSD, our proposed CRMSP achieved an improvement of f1-score that were 5.81\% and 1.20\% higher compared to FixMatch and the suboptimal model SimMatch based on LayoutLMv2, respectively.

This indicates that CRMSP can more effectively utilize labeled data to reduce model bias under long-tailed distribution. Furthermore, we observe that our method even achieves an f1-score of 91.30\%, which is close to the f1-score of 93.30\% achieved by the fully-supervised LayoutLMv3 on the KIE task, while using only 10\% of the labeled data.

\begin{table}
    \centering
    \resizebox{\textwidth}{!}{%
    \begin{tabular}{cccccccc}
    \hline
    Method                                                                          & Dataset    & \multicolumn{2}{c}{CIFAR10-LT} & \multicolumn{2}{c}{CIFAR100-LT} & \multicolumn{2}{c}{STL10-LT} \\
    & imb\_ratio & \multicolumn{2}{c}{$\gamma = \gamma _l = \gamma _u =100$}          & \multicolumn{2}{c}{$\gamma = \gamma _l = \gamma _u =10$}           & \multicolumn{2}{c}{$\gamma _l = 10, \gamma _u: unknown$}        \\ \cmidrule{2-8} 
    & \#Label    & $N_1=500$         & $N_1=1500$       & $N_1=50$          & $N_1=150$         & $N_1=150$        & $N_1=450$       \\
    & \#Unlabel  & $M_1=4000$        & $M_1=3000$       & $M_1=400$         & $M_1=300$         & $M_1=100k$       & $M_1=100k$      \\ \hline
    & Supervised & 46.75              & 62.78             & 31.11              & 49.02              & 45.39             & 62.09            \\ \hline
    \multirow{5}{*}{\begin{tabular}[c]{@{}c@{}}Classical \\ SSL\end{tabular}}             & FixMatch \cite{sohn2020fixmatch}   & 73.60              & 77.60             & \underline{48.52}              & 57.85              & 65.80             & 77.85            \\
    & Dash \cite{xu2021dash}      & 70.63              & 75.45             & 42.93              & 56.01              & 73.40             & 82.01            \\
    & FlexMatch \cite{zhang2021flexmatch}  & 62.18              & 73.69             & 38.99              & 53.88              & 82.75             & 85.55            \\
    & SimMatch \cite{zheng2022simmatch}       & \textbf{76.03}              & \underline{78.68}             & 47.30              & 57.86              & \underline{82.85}             & 85.75            \\
    & FreeMatch \cite{wang2022freematch}  & 70.63              & 76.41             & 44.16              & 57.20              & 82.34             & \underline{86.06}            \\
    \hline
    \multirow{4}{*}{\begin{tabular}[c]{@{}c@{}}Imbalanced\\ SSL\end{tabular}} & DARP \cite{kim2020distribution}      & \underline{75.92}              & 78.62             & \textbf{49.05}              & \underline{57.88}              & 65.38             & 75.70            \\
    & CReST \cite{wei2021crest}    & 71.21              & 75.74             & 44.37              & 55.77              & 63.58             & 71.70            \\
    & DASO \cite{oh2022daso}      & 71.12              & 76.79             & 48.35              & 57.71              & 68.37             & 79.03            \\
    & CRMSP(Ours)       & 71.38              & \textbf{78.91}             & 44.53              & \textbf{57.90}              & \textbf{83.01}             & \textbf{86.35}            \\ \hline
    \end{tabular}}
    \caption{Top-1 accuracy (\%) on CIFAR10-LT, CIFAR100-LT and STL10-LT. The best result is in \textbf{bold}, the second-best result is in \underline{underline}.
    \label{main-results3}
    }
\end{table}

\subsubsection{Results on CIFAR10/100-LT and STL10-LT}
To illustrate the generalization of CRMSP, we also conducted experiments on the CIFAR10/100-LT and STL10-LT, as shown in Table \ref{main-results3}. We consider rebalancing biased pseudo-labels by matching (e.g., $\gamma = \gamma_{l} = \gamma_{u}$) or mismatching (e.g., $\gamma_{l} = 10, \gamma_{u}: unknown$) distributions between imbalanced labeled and unlabeled data ($\mathcal{X}$ and $\mathcal{U}$) in Table \ref{main-results3}. When $\gamma_{l} = \gamma_{u}$, we compare the proposed CRMSP with several classical (i.e., FixMatch \cite{sohn2020fixmatch} and imbalanced (i.e., DARP \cite{kim2020distribution}, CReST \cite{wei2021crest} and DASO \cite{oh2022daso}) SSL baseline methods. In the supervised scenario, the performance is relatively constrained compared to other semi-supervised learning methods. 

Notably, CRMSP demonstrates comparable or even superior performance across most settings, exhibiting substantial improvements compared to the baseline methods SimMatch and DARP. Compared to the baseline FixMatch and baseline DARP, the accuracy has increased by 0.04\%-0.60\% and 0.02\%-17.63\%, respectively.

Overall, Table \ref{main-results3} demonstrates that our proposed CRMSP is not only effective in mitigating imbalance issues in the SSL domain of KIE but also applicable in the CV domain.

\begin{table}
    \centering
    \resizebox{\textwidth}{!}{%
    \begin{tabular}{cccccccccccc}
    \hline
    Method                          & Class      & 0             & 1             & 2             & 3             & 4             & 5             & 6             & 7             & 8             & 9             \\ \hline
                                    & Supervised & 86.1          & 61.0          & 81.5          & 49.4          & 38.8          & 23.8          & 42.9          & 18.5          & 29.0          & 23.0          \\ \cmidrule(r){2-12}
    \multirow{4}{*}{Classical SSL}    & FixMatch \cite{sohn2020fixmatch}   & 73.8          & 4.7           & 0.2           & 48.2          & 80.4          & 53.9          & 0.2           & 1.1           & 42.1          & 38.1          \\
                                    & Dash \cite{xu2021dash}       & 
                                    94.1    & \textbf{92.0} & \textbf{96.1} & \textbf{60.0} & 86.3          & 70.5    & 83.8          & 50.0          & 69.0          & 32.3          \\
                                    & FlexMatch \cite{zhang2021flexmatch}  & \textbf{95.4}  & 83.1 &  \underline{94.6}         & 54.4   & 85.8  & 69.9  & 89.6    & \textbf{80.3} &  94.0  & 80.5    \\
                                    & SimMatch \cite{zheng2022simmatch}  & 92.0              & \underline{85.0}              & 94.4              &  58.0             & 87.9              &  \underline{73.0}             &  \underline{90.4}             & 69.0              & 95.0              & \underline{84.3}              \\ 
                                    & FreeMatch \cite{wang2022freematch}  & \underline{94.5} & 84.3    & 94.1          & 46.3          & \textbf{90.3} & 67.8          & 88.9    & \underline{77.6} & \textbf{95.8} &  84.0    \\ \hline
    \multirow{4}{*}{Imbalanced SSL} & DARP \cite{kim2020distribution}      & 93.3          & 82.6          & 93.3          & 54.4          & 81.0          & 34.9          & 79.9          & 26.4          & 64.5          & 43.6          \\
                                    & CReST \cite{wei2021crest}     & 55.3          & 37.3          & 58.6          & \underline{59.5}    & 68.4          & 20.9          & 73.5          & 53.1          & 78.6          & 82.1          \\
                                    & DASO \cite{oh2022daso}      & 90.8          & 75.3          & 92.3          & 52.9          & 82.6          & 53.8          & 82.8          & 31.0          & 72.1          & 50.4          \\
                                    & CRMSP(Ours)       & 93.0          & 82.6          &  94.3    & 56.0          & \underline{89.0}    & \textbf{73.8} & \textbf{90.5} & 68.9    & \underline{95.0}    & \textbf{87.1} \\ \hline
    \end{tabular}}
    \caption{Per-class top-1 accuracy (\%) on the balanced test dataset of STL10-LT ($\gamma_{l}=10,\gamma_{u}:unknown,N_{1}=150,M_{1}=100k$). Our method shows a significant improvement in pseudo-labeling for tail classes. The best result is in \textbf{bold}, the second-best result is in \underline{underline}.}
    \label{per-class}
\end{table}

\subsubsection{Per-class performance}
By comparing the top-1 accuracy of different methods on the STL10-LT across per class, we designate classes C0-C4 as the head classes and C5-C9 as the tail classes. In the head classes, it can be observed that Dash performs well on some head classes (C1, C2, C3), while our method achieves the second-best accuracy in C4 with 89.0\%. In the tail classes, our method achieves an accuracy of 95.0\% for C8, which is second-best compared to FreeMatch. Remarkably, our method CRMSP achieves the best accuracy of 73.8\%, 90.5\%, and 87.1\% in C5, C6, and C9, respectively, representing improvements of 0.8\%, 0.1\%, and 2.8\% over the second-best class SimMatch. This table highlights the enhancement provided by our method for tail classes in long-tailed distribution.

\subsection{Ablation study}
To verify the effectiveness of each component of our proposed method, We conduct extensive ablation studies on the CORD. The results are shown in Table \ref{table3}. And the evaluation metric for all experiments was the f1-score. We utilize LayoutLMv3 as the base model. 

\begin{table}
    \centering
    \resizebox{0.5\textwidth}{!}{%
    \begin{tabular}{cccccc}
    \hline
    {\#} & {\textbf{RP}} & {\textbf{$L_{ctr}$}} & {\textbf{MP}} & {\textbf{FUNSD}} & {\textbf{CORD}} \\
    \hline
    {0} & {} & {} & {} & {$66.56$} &{$84.05$} \\
    {1} & {} & {$\checkmark$} & {} & {$67.21$} & {$84.76$} \\
    {2} & {} & {$\checkmark$} & {$\checkmark$} & {$67.75$} & {$85.36$} \\
    {3} & {$\checkmark$} & {} & {} & {$68.96$} & {$87.56$} \\
    {4} & {$\checkmark$} & {$\checkmark$} & {} & {$69.91$} & {$88.08$} \\
    {5} & {$\checkmark$} & {$\checkmark$} & {$\checkmark$}  & {$71.12$} & {$89.59$} \\
    \hline
    \end{tabular}
    }
    \caption{Ablation study on FUNSD and CORD. 
    “RP”, “$L_{ctr}$” and “MP” mean Reweighting Pseudo-Labels, Contrastive Loss and Merged Prototypes, respectively. We use supervised loss as baseline.}
    \label{table3}
\end{table}

\subsubsection{Effectiveness of Reweighting Pseudo-Labels}
To verify the effectiveness of RP, By comparing Experiment 2 and Experiment 5 in Table \ref{table3}, we can observe that the RP improves the f1-score by 3.37\% and 4.23\% on the FUNSD and CORD. Table \ref{table4} We also compared the performance of tail classes without and with RP. It is found that by adding RP, there is a significant improvement in the results of the tail classes, especially for $num.sub\_cnt$ (0.42→0.78). Note that LayoutLMv2 is used as the base model.

To demonstrate the detailed effectiveness of our proposed RP, we present confusion matrixes of the predictions on the test dataset of FUNSD. As depicted in Fig. \ref{confusion_matrixes}, the pseudo-labels for the tail classes without RP (e.g., C4, C5 and C6) are underestimated, while the accuracy between the pseudo-labels and the true labels for the head classes is higher. Our proposed RP improves the generation of more balanced pseudo-labels for tail classes, alleviating the issue of long-tailed distribution.

\begin{table}
    \centering
    \resizebox{\textwidth}{!}{%
    \begin{tabular}{cccccccc}
    \hline
    Class   & menu.sub\_price & menu.sub\_cnt & total.creditcardprice & sub\_total.service\_price & menu.num      & sub\_total.etc & sub\_total.discount\_price \\ \hline
    support & {$20$}              & {$17$}            & {$16$}                    & {$12$}                        & {$11$}            & {$8$}              & {$7$}                          \\ \hline
    w/o RP  & {$0.72$}            & {$0.42$}          & {$0.46$}                  & {$0.88$}                      & {$0.40$}          & {$0.33$}           & {$0.73$}                       \\
    w RP    & {$\mathbf{0.88}$}   & {$\mathbf{0.78}$} & {$\mathbf{0.73}$}  & {$\mathbf{0.96}$}   & {$\mathbf{0.74}$} & {$\mathbf{0.62}$}  & {$\mathbf{0.88}$}             \\ \hline
    \end{tabular}}
    \caption{The influence of the RP on the f1-score of tail classes on the CORD, "support" denotes the sample size.}
    \label{table4}
\end{table}

\subsubsection{Effectiveness of Contrastive Loss} When incorporating contrastive loss, CRMSP can further boost the performance on all settings by another few points, resulting in 0.52\% to 0.66\% absolute accuracy improvement by comparing Experiment 3 and Experiment 4 in Table \ref{table3}.

\subsubsection{Effectiveness of Merged Prototypes}
Comparing Experiment 4 and Experiment 5 in Table \ref{table3}, we observed that f1-score improved by 1.21\% and 1.51\% on the FUNSD and CORD, respectively.

\begin{figure}[htbp]
  \centering
  \begin{minipage}[b]{0.48\linewidth}
    \centering
    \includegraphics[width=\linewidth, trim=1 1 4 1, clip]{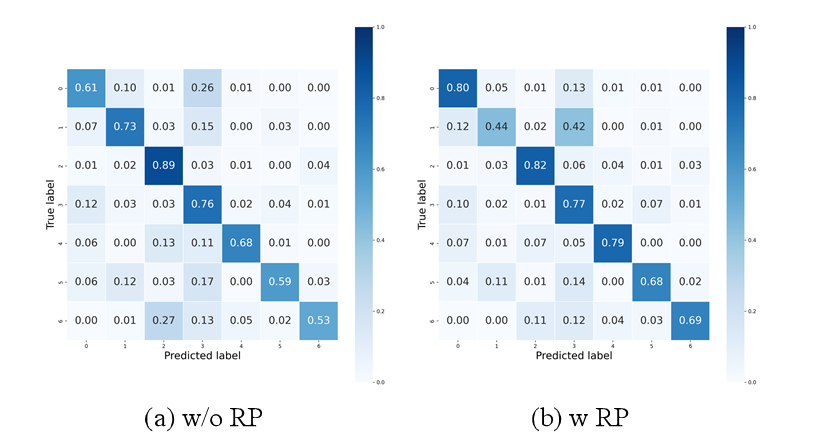}
    \captionof{figure}{Confusion matrixes of the predictions on the test dataset of FUNSD.}    
    \label{confusion_matrixes}
  \end{minipage}%
  \hspace{0.5cm}
  \begin{minipage}[b]{0.48\linewidth}
    \centering
    \includegraphics[width=\linewidth, trim=10 2 3 1, clip]{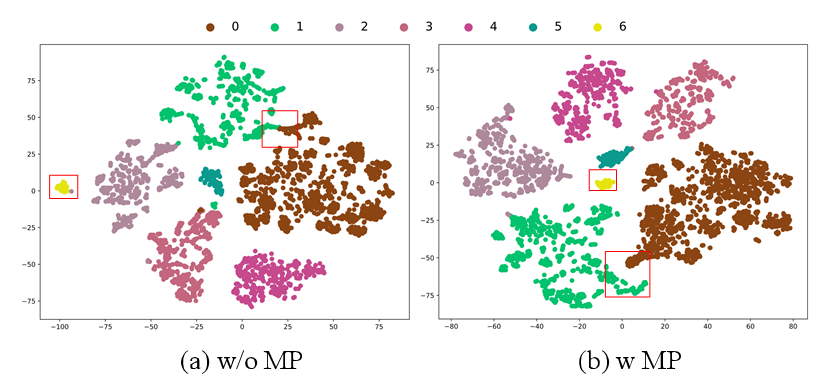}
    \captionof{figure}{Comparison of t-SNE visualization of unlabeled data on the FUNSD.}
    \label{figure4}
  \end{minipage}
\end{figure}

To demonstrate the effectiveness of our proposed MP, we present the comparison of t-SNE visualization of unlabeled data. As shown in Fig. \ref{figure4}, the MP helps the tail class (e.g., C6) to be separated from the confusion class and better clustering is achieved. Other confusing features (e.g., C0 and C1)  are also clustered more compactly. Fig. \ref{figure4} effectively promotes intra-class compactness and inter-class separability of unlabeled tail classes in feature space.

\subsubsection{Ablation study on $\lambda_{un}$} In Fig. \ref{fig:img1}, we study the effect of the temperature hyper-parameter $\lambda_{un}$ to compute the weights for unsupervised loss described in Eq. \ref{total-loss}. We empirically find that, for both FUNSD and CORD, $\lambda_{un}=0.1$ shows the best performance.


\subsubsection{Ablation study on $\lambda_{ctr}$} In Fig. \ref{fig:img2}, we investigate the impact of the temperature hyper-parameter $\lambda_{ctr}$ on computing the weights for the contrastive loss described in Eq. \ref{total-loss}. $\lambda_{ctr}=0.1$ yields the optimal performance on the FUNSD and CORD.


\subsubsection{Ablation study on $K$} 
\begin{figure}[htbp]
  \centering
  \begin{minipage}[b]{0.32\linewidth}
    \centering
    \includegraphics[width=\linewidth]{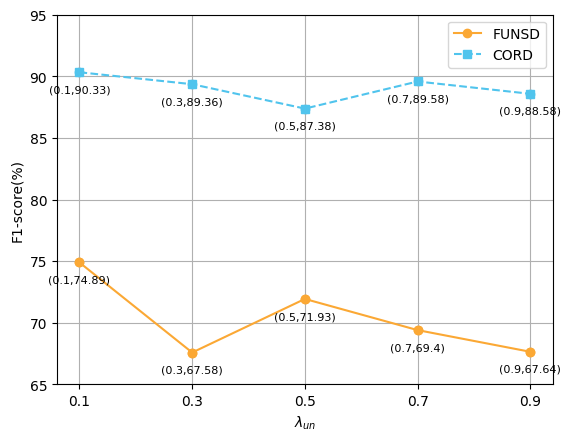}
    \captionof{figure}{The influence of different $\lambda_{un}$ on the f1-score on the FUNSD and CORD.}
    \label{fig:img1}
  \end{minipage}%
  \hspace{0.2cm}
  \begin{minipage}[b]{0.32\linewidth}
    \centering
    \includegraphics[width=\linewidth]{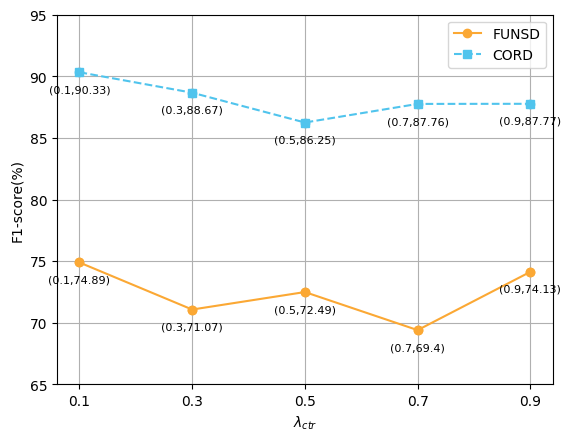}
    \captionof{figure}{The influence of different $\lambda_{ctr}$ on the f1-score on the FUNSD and CORD.}
    \label{fig:img2}
  \end{minipage}%
  \hspace{0.2cm}
  \begin{minipage}[b]{0.32\linewidth}
    \centering
    \includegraphics[width=\linewidth]{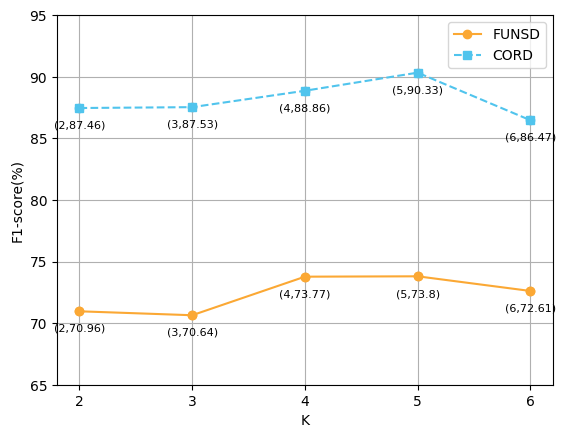}
    \captionof{figure}{The influence of different $K$ on the f1-score on the FUNSD and CORD.}
    \label{fig:img3}
  \end{minipage}
\end{figure}

The influence of different $K$ on the f1-score on the FUNSD and CORD is illustrated in Fig. \ref{fig:img3}. We notice that $K=5$ provides the best f1-score among all tested values. When $K$ is set to a small value, prototypes for some tail samples lack representation. Comparing these tail samples with non-representative prototypes results in semantic pseudo-labels that push the feature in the wrong direction in the feature space, leading to classification errors. On the other hand, if $K$ is set too large, although the new sample features are effectively separated from classes not belonging to this super-class, the super feature range extends far beyond the range of variations in tail features, causing internal confusion within the super-class.



\subsubsection{Case Study} 
\begin{figure}
\begin{flushright}
\begin{minipage}[b]{0.3\linewidth}
  \centering
·  \centerline{\includegraphics[width=0.95\textwidth]{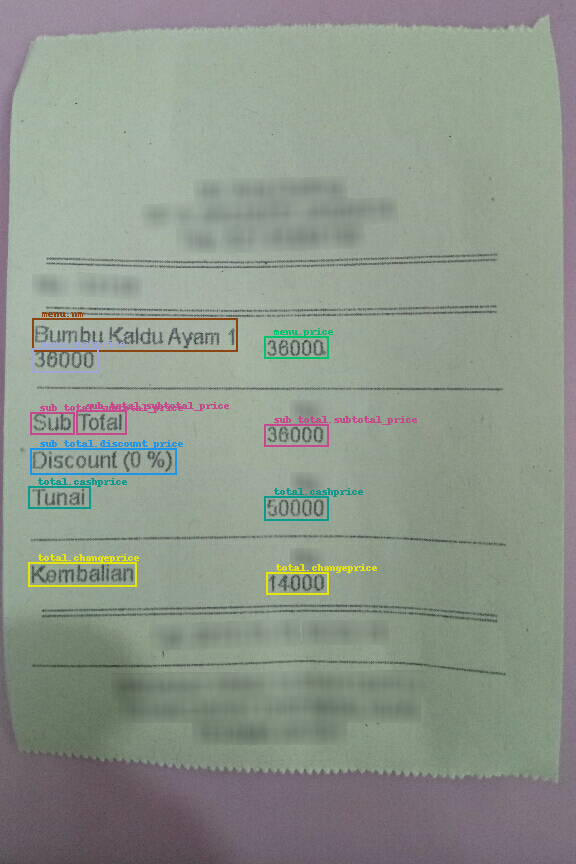}}
\end{minipage}
\begin{minipage}[b]{0.3\linewidth}
  \centering
  \centerline{\includegraphics[width=0.95\textwidth]{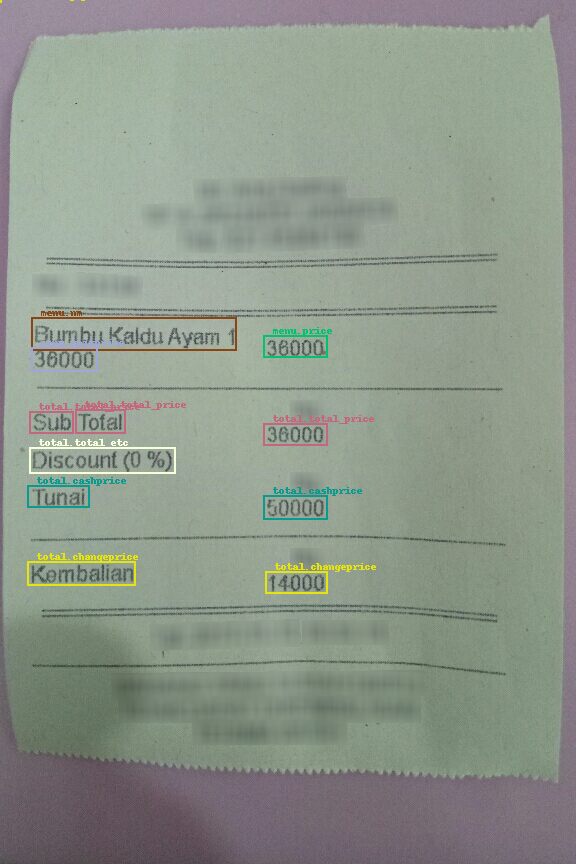}}
\end{minipage}
\begin{minipage}[b]{0.3\linewidth}
  \centering
  \centerline{\includegraphics[width=0.95\textwidth]{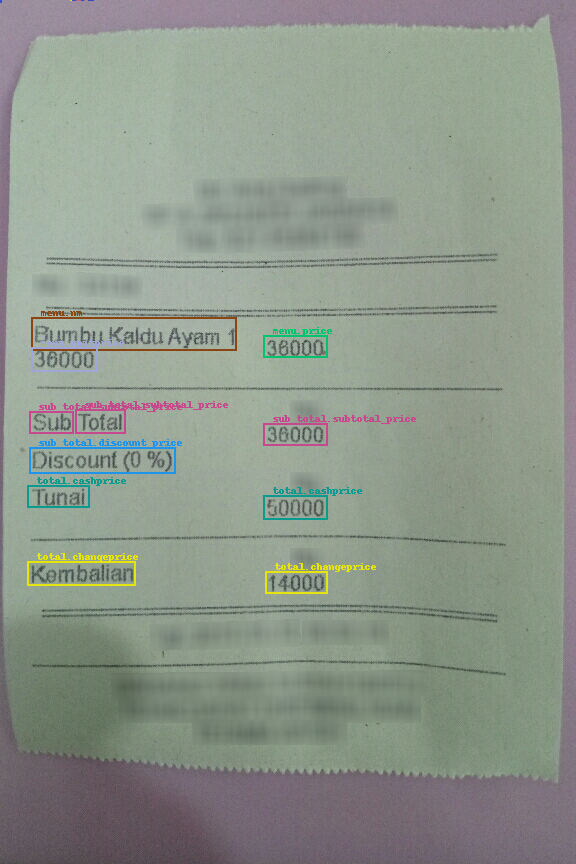}}
\end{minipage}
\begin{minipage}[b]{0.3\linewidth}
  \centering
  \centerline{\includegraphics[width=0.95\textwidth]{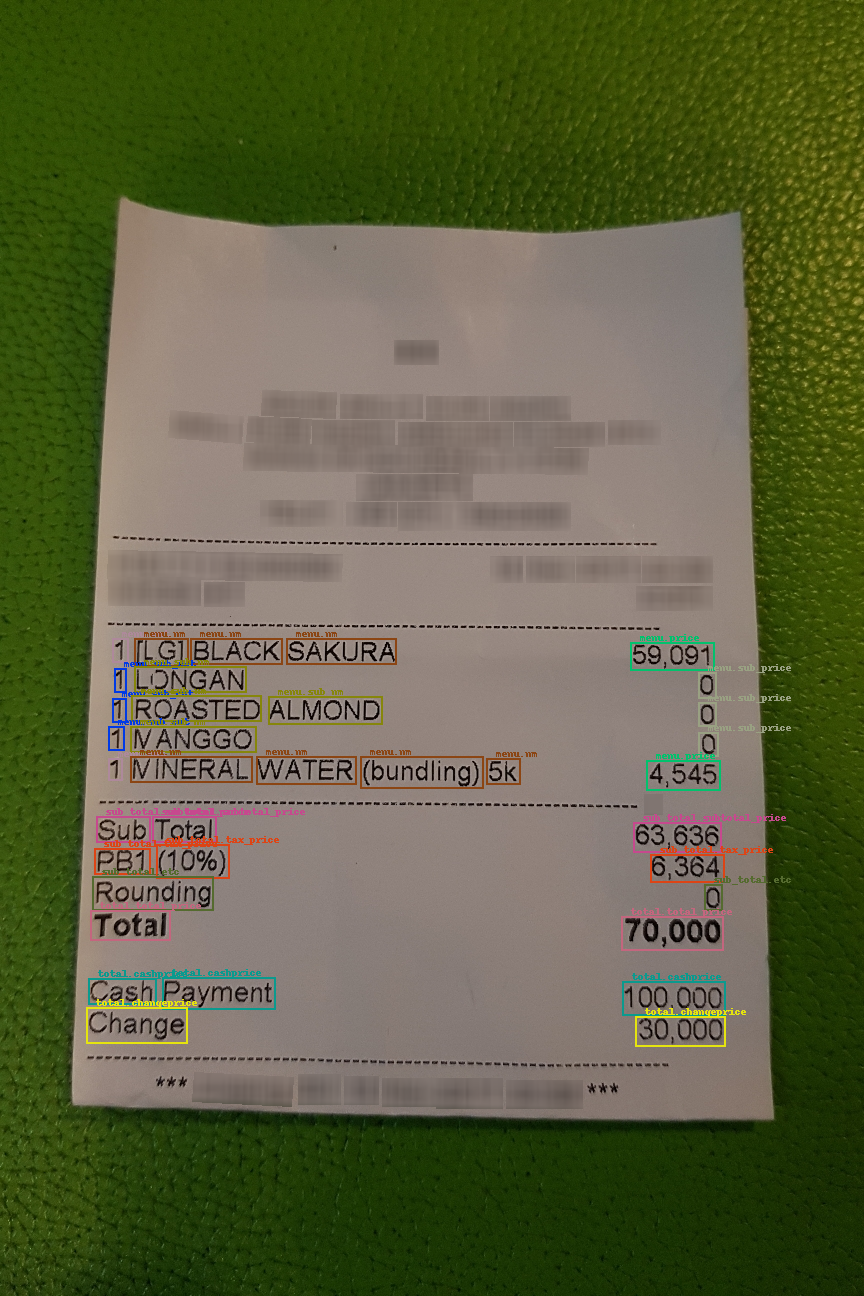}}
\end{minipage}
\begin{minipage}[b]{0.3\linewidth}
  \centering
  \centerline{\includegraphics[width=0.95\textwidth]{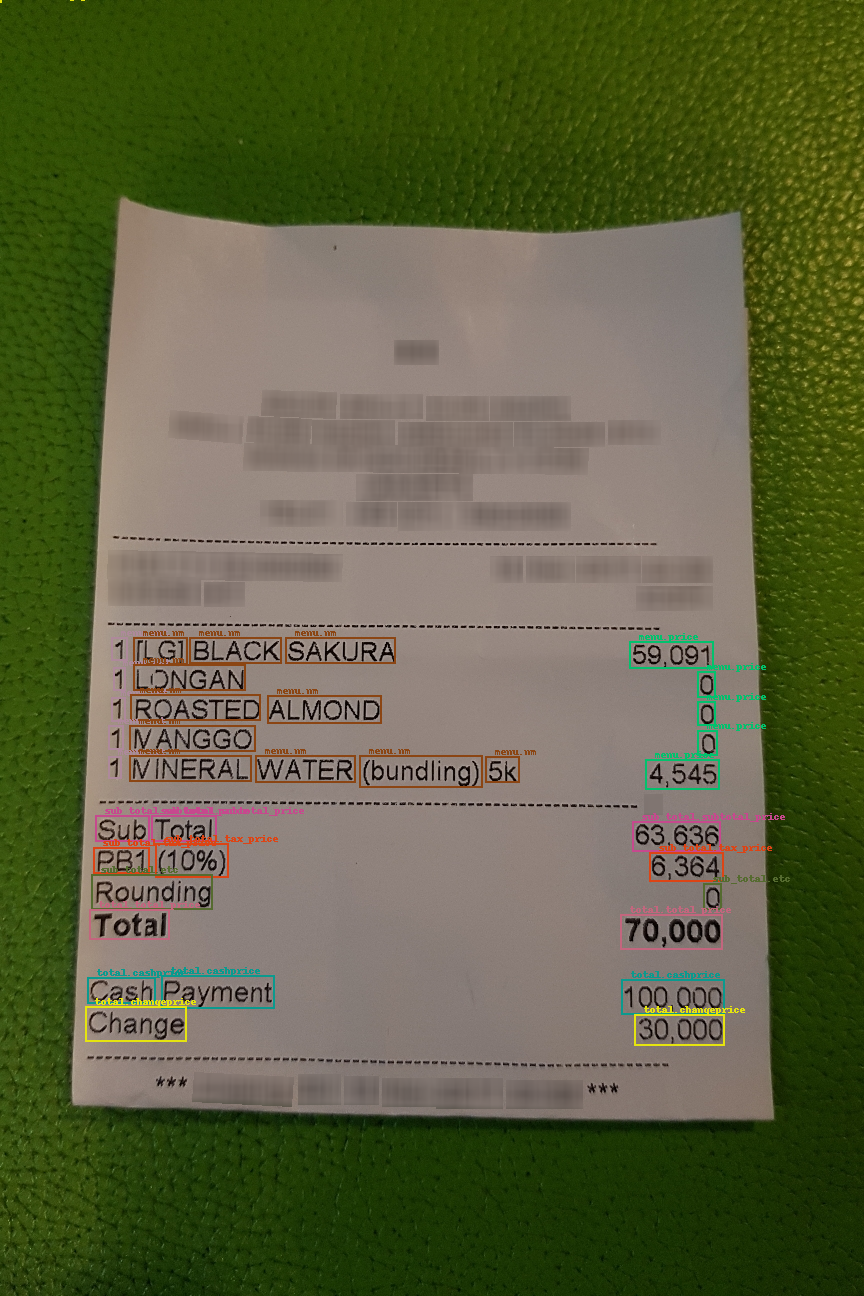}}
\end{minipage}
\begin{minipage}[b]{0.3\linewidth}
  \centering
  \centerline{\includegraphics[width=0.95\textwidth]{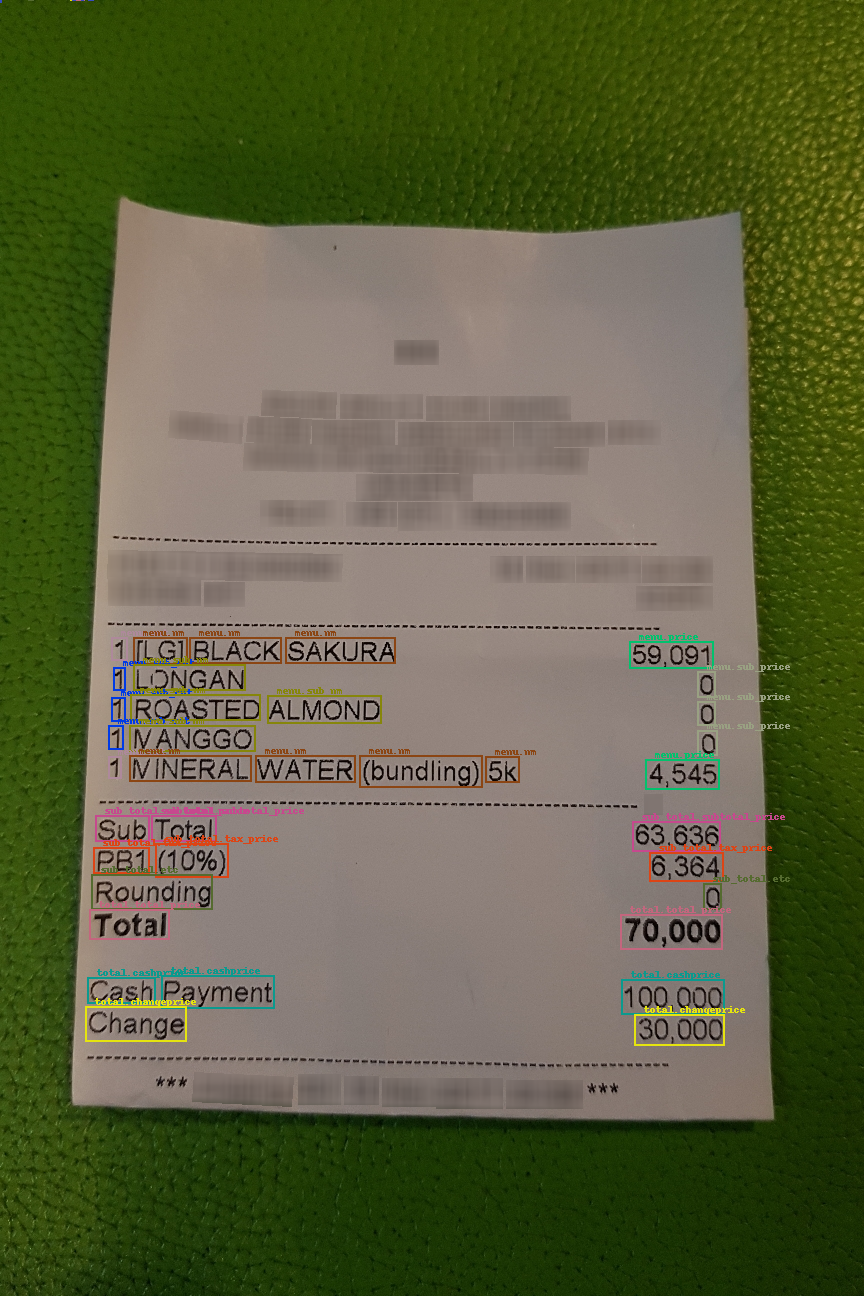}}
\end{minipage}
\begin{minipage}[b]{0.3\linewidth}
  \centering
  \centerline{\includegraphics[width=0.95\textwidth]{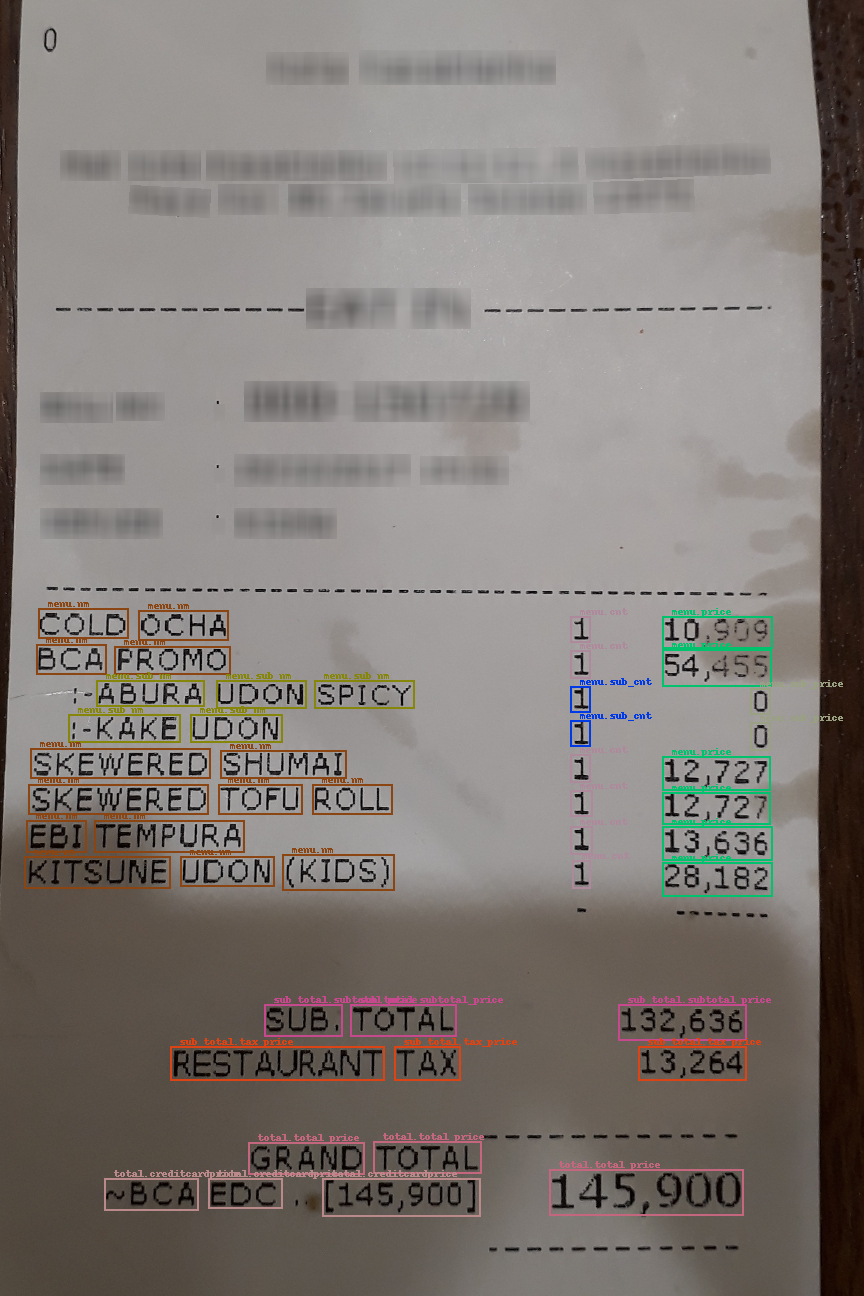}}
  \centerline{(a) Ground-truth}\medskip
\end{minipage}
\begin{minipage}[b]{0.3\linewidth}
  \centering
  \centerline{\includegraphics[width=0.95\textwidth]{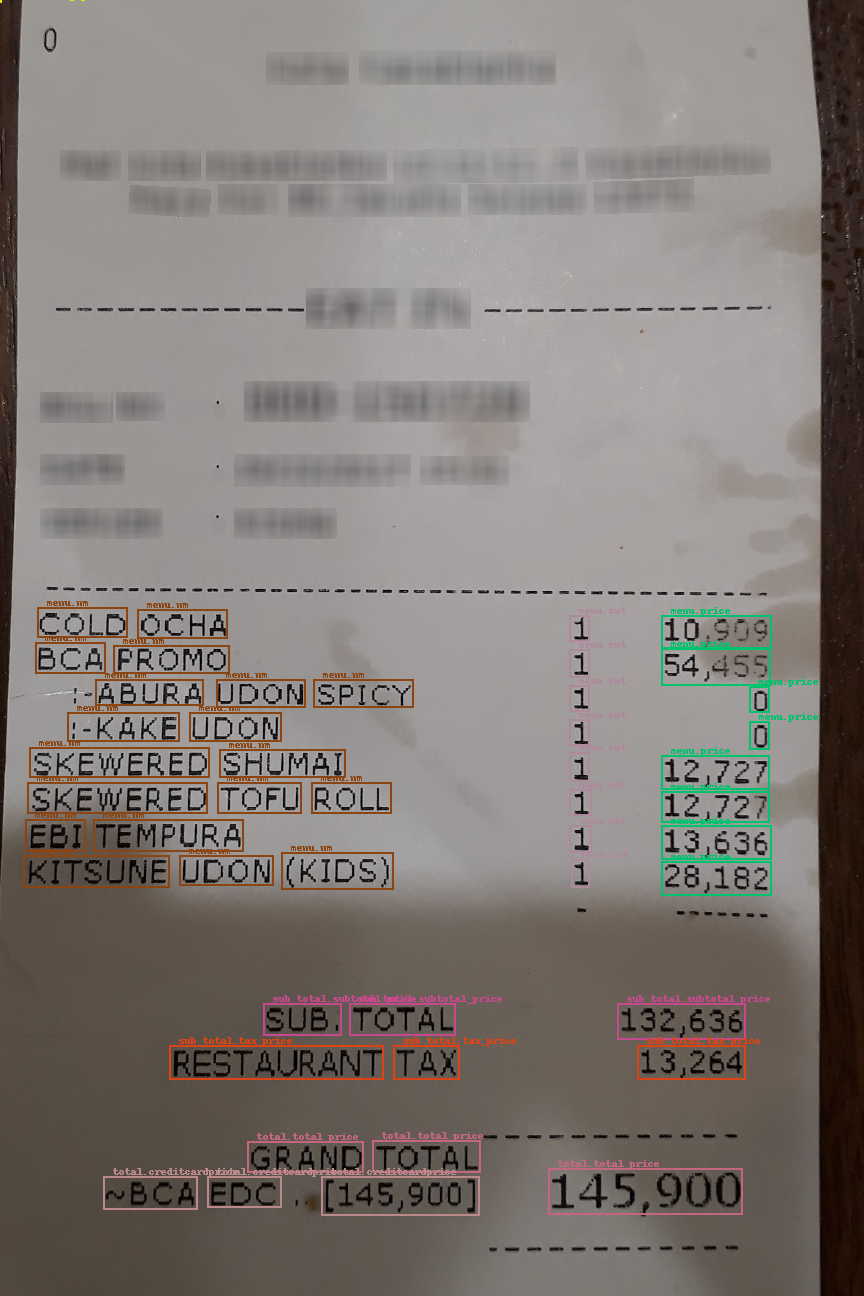}}
  \centerline{(b) FixMatch}\medskip
\end{minipage}
\begin{minipage}[b]{0.3\linewidth}
  \centering
  \centerline{\includegraphics[width=0.95\textwidth]{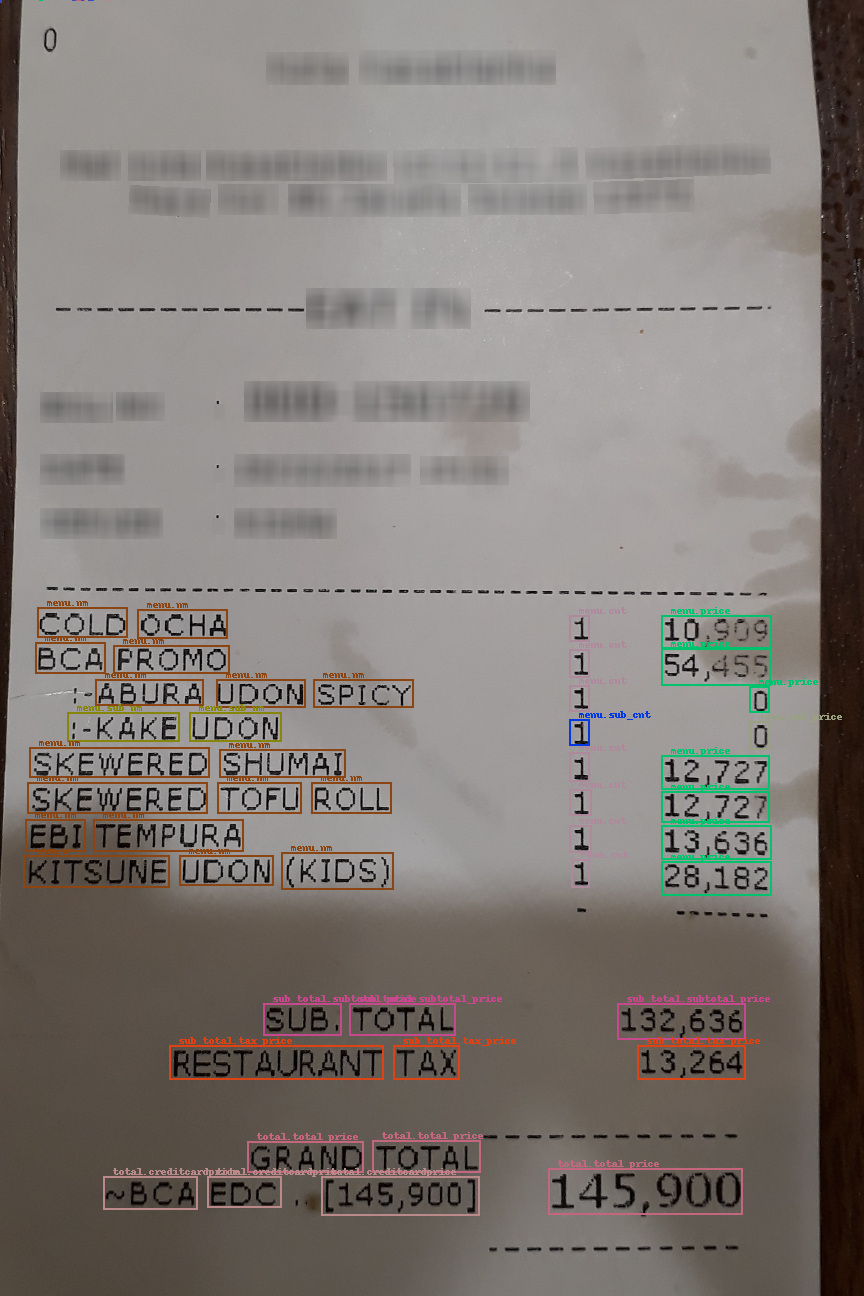}}
  \centerline{(c) CRMSP}\medskip
\end{minipage}
\end{flushright}
\caption{Example output of Ground-truth, FixMatch, and CRMSP for CORD.}
\label{fig3}
\end{figure}

We present the output of samples for Ground-truth, FixMatch, and CRMSP on both the CORD and FUNSD. On the CORD, as depicted in Fig. \ref{fig3}, the tail class $sub\_total.discount\_price$ in the ground-truth is incorrectly classified as $total.total\_etc$ by FixMatch. This misclassification is corrected by our proposed CRMSP approach. The tail classes $menu.sub\_cnt$ and $menu.sub\_nm$ are erroneously associated with their respective classes $menu.cnt$ and $menu.nm$, but our proposed CRMSP method adeptly distinguishes between them.

\begin{figure}[t]
\begin{flushright}
\begin{minipage}[b]{0.32\linewidth}
  \centering
  \centerline{\includegraphics[width=0.95\textwidth]{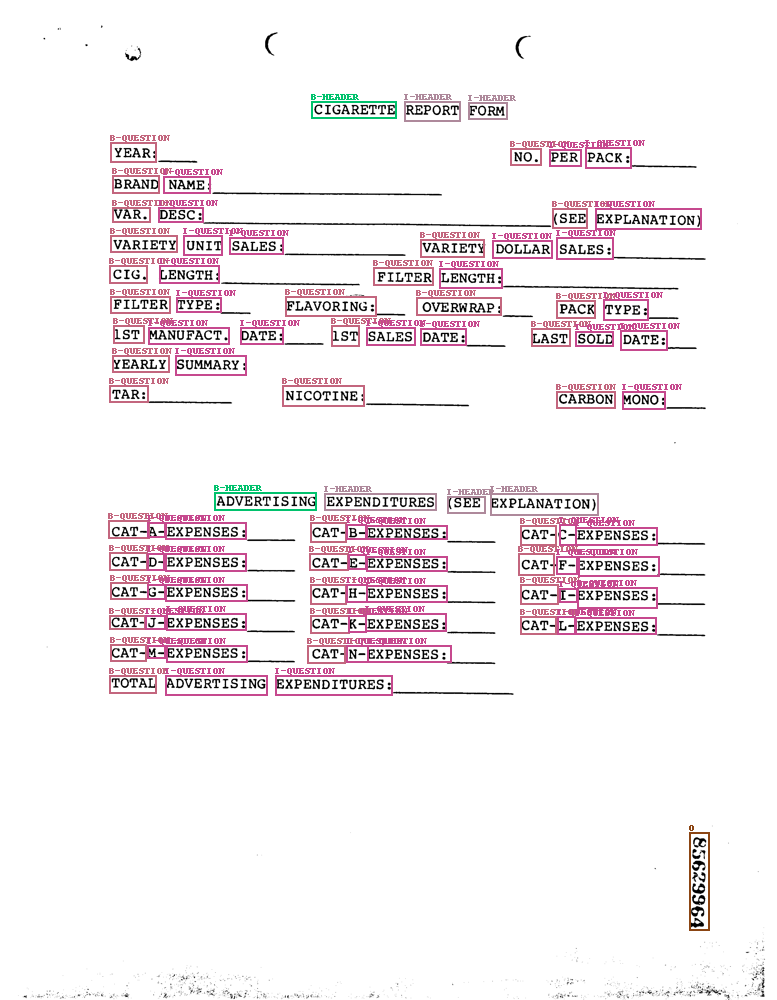}}
\end{minipage}
\begin{minipage}[b]{0.32\linewidth}
  \centering
  \centerline{\includegraphics[width=0.95\textwidth]{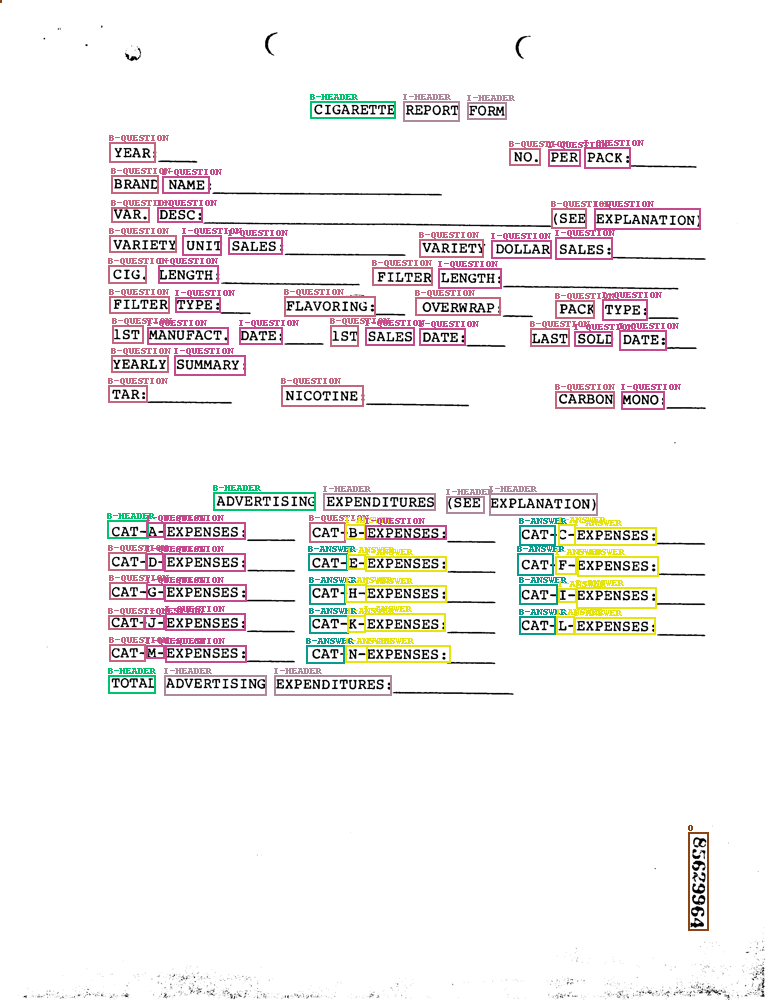}}
\end{minipage}
\begin{minipage}[b]{0.32\linewidth}
  \centering
  \centerline{\includegraphics[width=0.95\textwidth]{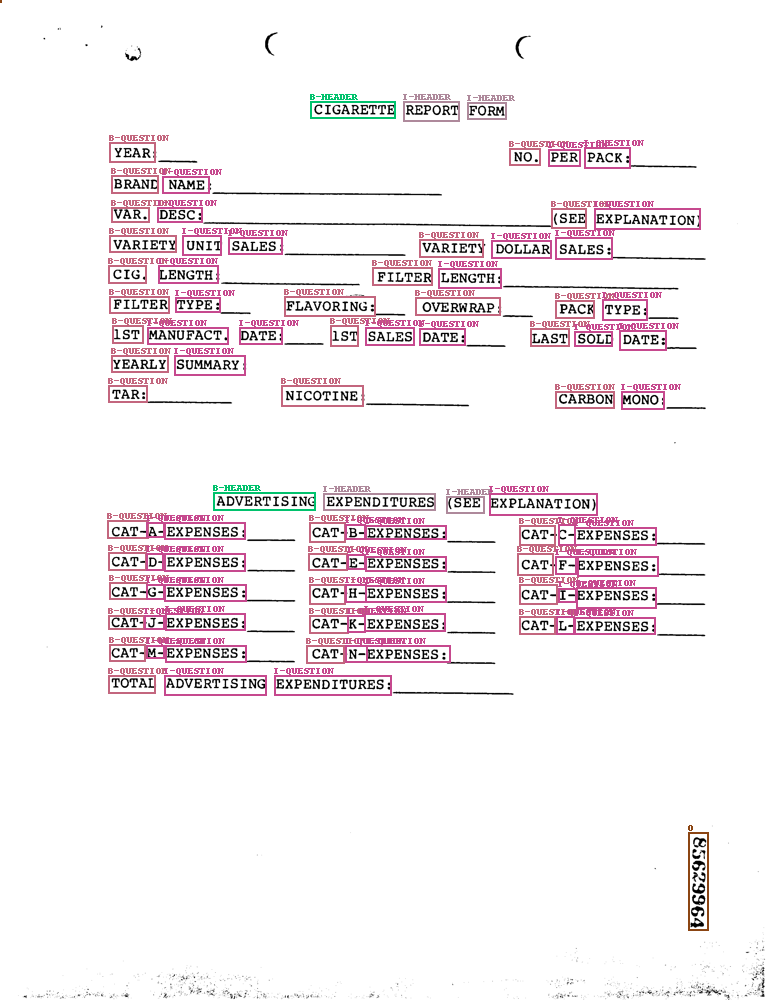}}
\end{minipage}
\begin{minipage}[b]{0.32\linewidth}
  \centering
  \centerline{\includegraphics[width=0.95\textwidth]{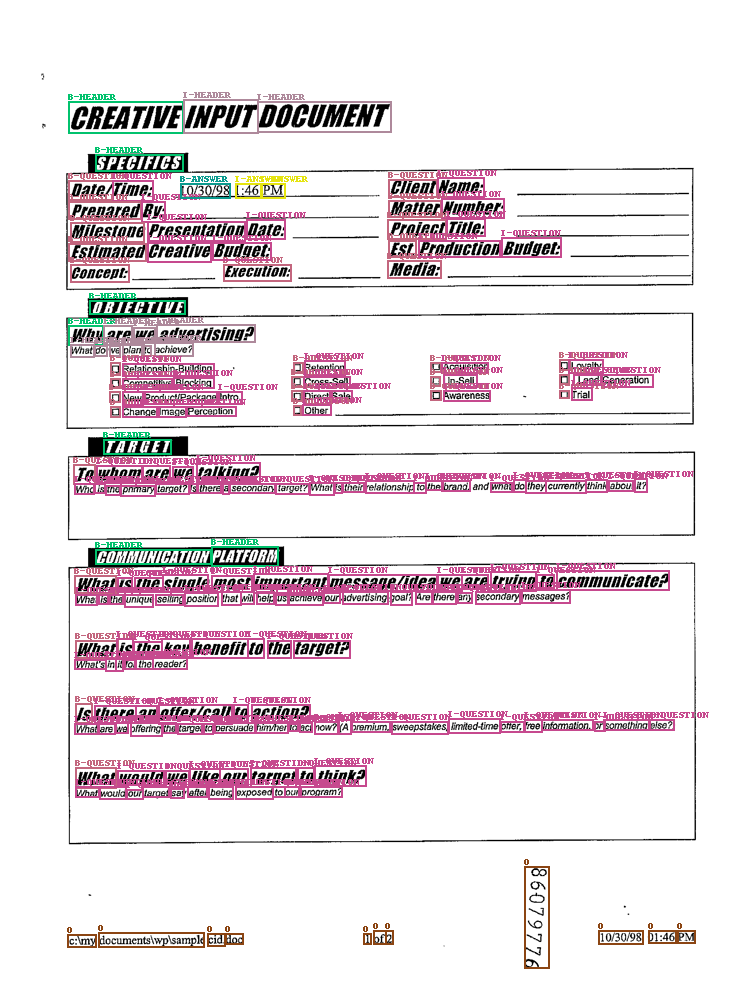}}
  \centerline{(a) Ground-truth}\medskip
\end{minipage}
\begin{minipage}[b]{0.32\linewidth}
  \centering
  \centerline{\includegraphics[width=0.95\textwidth]{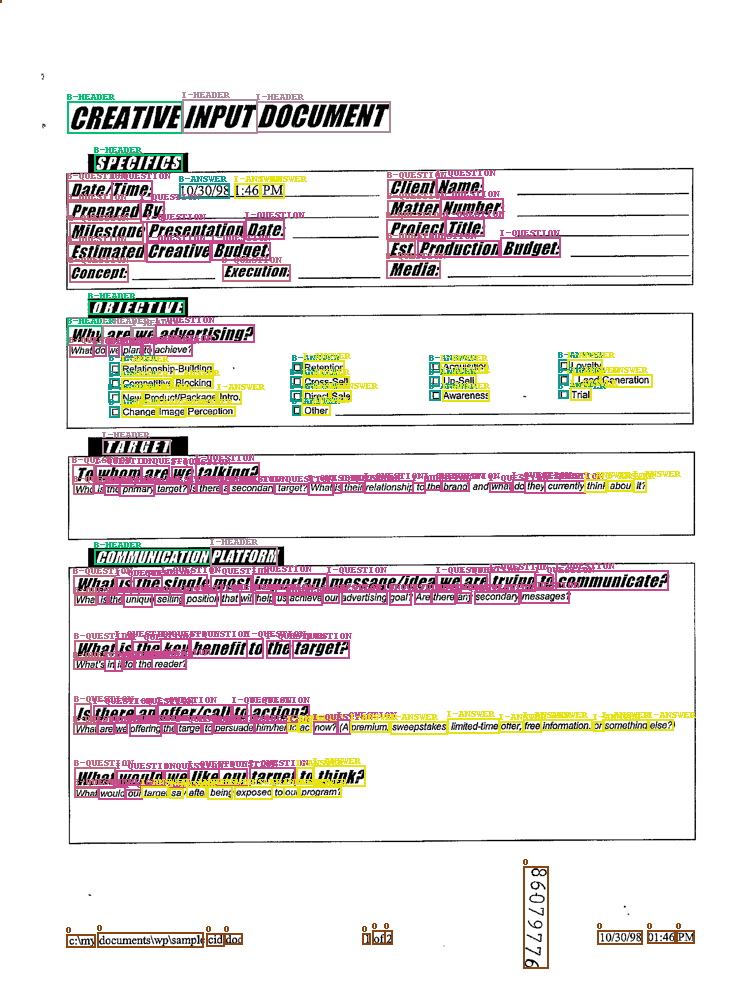}}
  \centerline{(b) FixMatch}\medskip
\end{minipage}
\begin{minipage}[b]{0.32\linewidth}
  \centering
  \centerline{\includegraphics[width=0.95\textwidth]{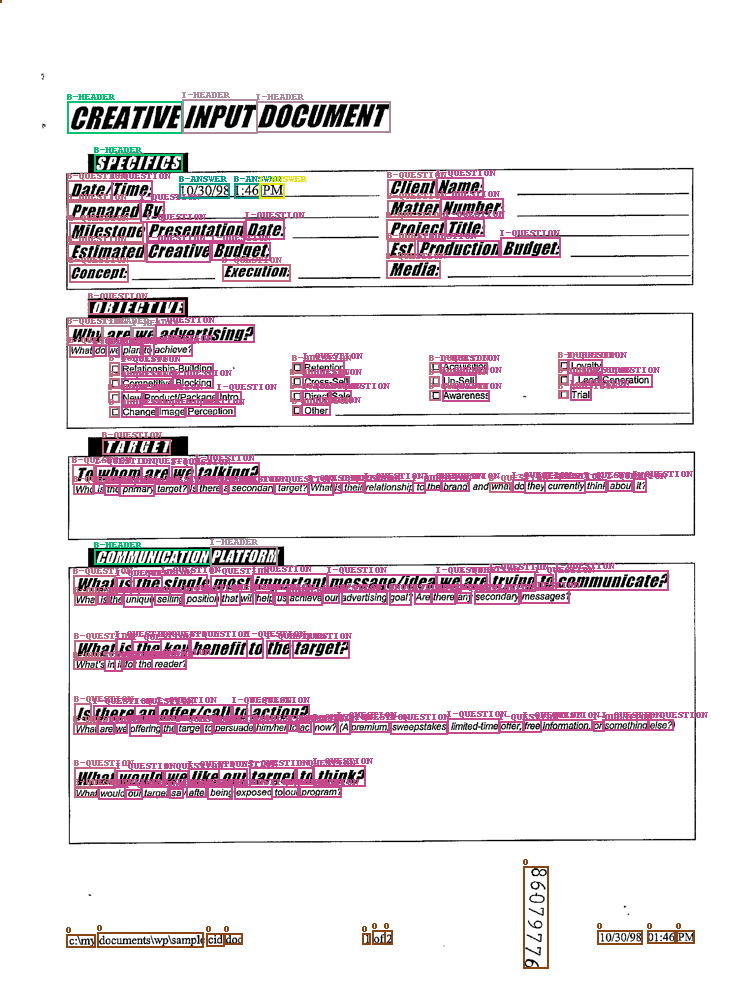}}
  \centerline{(c) CRMSP}\medskip
\end{minipage}

\end{flushright}
\caption{Example output of Ground-truth, FixMatch, and CRMSP for FUNSD.}
\label{fig11}
\end{figure}

On the FUNSD, as shown in Fig. \ref{fig11}, FixMatch misclassifies B-QUESTION and I-QUESTION as B-ANSWER and I-ANSWER, respectively. However, CRMSP correctly identifies these tail classes.

\section{Conclusion}
\label{conclusion}

In this paper, we propose a novel semi-supervised approach for key information extraction with Class-Rebalancing and Merged Semantic Pseudo-Labeling (CRMSP). Firstly, the Class-Rebalancing Pseudo-Labeling (CRP) module is proposed to directly rebalance pseudo-labels with a reweighting factor, increasing attention to tail classes. Secondly, the Merged Semantic Pseudo-Labeling (MSP) module is proposed to achieve intra-class compactness and inter-class separability of unlabeled tail classes in feature space by assigning samples to Merged Prototypes (MP). We even achieved close to fully-supervised learning in the semi-supervised setting. Extensive experimental results have demonstrated the proposed CRMSP surpasses other state-of-the-art methods on three benchmarks. Our findings suggest that the proposed approach can obtain high-quality pseudo-labels from a larger amount of unlabeled data, which provides a good solution for semi-supervised learning.



 \bibliographystyle{elsarticle-num} 
 \bibliography{sn-bibliography}





\end{document}